%% file: main.tex
\documentclass[10pt,twocolumn,letterpaper]{article}

\usepackage[pagenumbers]{cvpr} 
\usepackage{fixltx2e}

\input{preamble}

\definecolor{cvprblue}{rgb}{0.21,0.49,0.74}
\usepackage[pagebackref,breaklinks,colorlinks,allcolors=cvprblue]{hyperref}

\title{D³-Predictor: Noise-Free Deterministic Diffusion for Dense Prediction}

\author{Changliang Xia\thanks{Equal contribution} \and
Chengyou Jia\footnotemark[1] \and
Minnan Luo\thanks{Corresponding author} \and
Zhuohang Dang \and
Xin Shen \and
Bowen Ping\\
School of Computer Science and Technology, Xi'an Jiaotong University\\
{\tt\small 202066@stu.xjtu.edu.cn, cp3jia@stu.xjtu.edu.cn, minnluo@xjtu.edu.cn,} \\
{\tt\small dangzhuohang@stu.xjtu.edu.cn, 2213311110@stu.xjtu.edu.cn, jayceping6@gmail.com}
}

\begin{document}
\twocolumn[{
\renewcommand\twocolumn[1][]{#1}
\maketitle
\begin{center}
    \vspace{-2em}
    \includegraphics[width=0.85\textwidth]{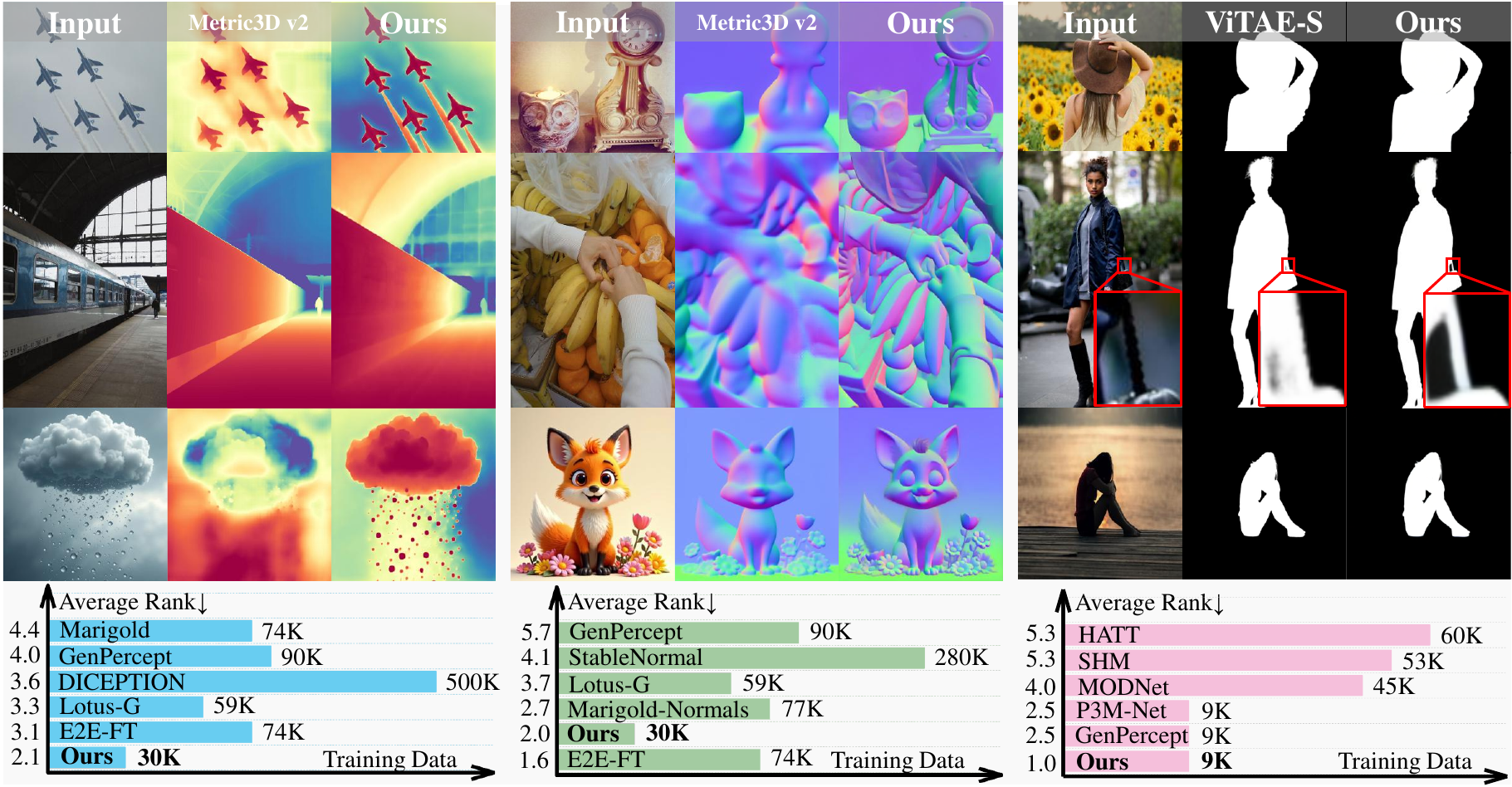}
    \captionsetup{type=figure}
    \caption{%
    We present D³-Predictor, a noise-free deterministic diffusion model achieving superior performance and generalization across various dense prediction tasks, with less than half the training data previously used and efficiently performing inference in a single step.
    }
    \label{fig:teaser}
\end{center}
}]

\def\thefootnote{*}\footnotetext{Equal Contribution}
\def\thefootnote{†}\footnotetext{Corresponding Author}
\input{sec/0_abstract}    
\input{sec/1_intro}
\input{sec/2_related_work}
\input{sec/3_method}
\input{sec/4_experiment}
\input{sec/5_conclusion}
{
    \small
    \bibliographystyle{ieeenat_fullname}
    \bibliography{main}
}

\input{sec/X_suppl}
\end{document}

%% file: preamble.tex
\usepackage{multirow}
\usepackage{colortbl}
\usepackage{array}
\usepackage{tabularx}
\usepackage{makecell}
\usepackage{adjustbox}
\usepackage{booktabs}
\usepackage{xcolor}
\usepackage{tcolorbox}
\newcolumntype{M}[1]{>{\centering\arraybackslash}m{#1}}



%% file: sec/0_abstract.tex
\begin{abstract}

Although diffusion models with strong visual priors have emerged as powerful dense prediction backbones, they overlook a core limitation: the stochastic noise at the core of diffusion sampling is inherently misaligned with dense prediction that requires a deterministic mapping from image to geometry. In this paper, we show that this stochastic noise corrupts fine-grained spatial cues and pushes the model toward timestep-specific noise objectives, consequently destroying meaningful geometric structure mappings. To address this, we introduce \textbf{D³-Predictor}, a noise-free deterministic diffusion-based dense prediction model built by reformulating a pretrained diffusion model without stochasticity noise. Instead of relying on noisy inputs to leverage diffusion priors, D³-Predictor views the pretrained diffusion network as an ensemble of timestep-dependent visual experts and self-supervisedly aggregates their heterogeneous priors into a single, clean, and complete geometric prior. Meanwhile, we utilize task-specific supervision to seamlessly adapt this noise-free prior to dense prediction tasks. Extensive experiments on various dense prediction tasks demonstrate that D³-Predictor achieves competitive or state-of-the-art performance in diverse scenarios. In addition, it requires less than half the training data previously used and efficiently performs inference in a single step. Our code, data, and checkpoints are publicly available at \url{https://x-gengroup.github.io/HomePage_D3-Predictor/}.
\end{abstract}

%% file: sec/1_intro.tex
\vspace{-1.5em}

\section{Introduction}
\label{sec:intro}

Dense prediction, such as depth estimation~\cite{ming2021deep} and surface normal estimation~\cite{wang2015designing}, are fundamental tasks in computer vision, with numerous applications like autonomous driving~\cite{cabon2020virtual, godard2019digging}, scene reconstruction~\cite{chen2024nc, zhang2025monoinstance}, inverse rendering~\cite{chen2024intrinsicanything, zeng2024rgbx}, and so on. Although state-of-the-art discriminative dense prediction models~\cite{bochkovskii2024depth, hu2024metric3d} achieve impressive performance, they still struggle to capture fine-grained high-frequency details. To address this limitation, current works~\cite{ke2024repurposing, lee2024exploiting, fu2024geowizard, ye2024stablenormal, zhao2025diception} reformulate dense prediction as an image-conditioned iterative denoising process based on diffusion models~\cite{rombach2022high}. By leveraging powerful visual priors of diffusion models, these methods can produce dense predictions results with fine-grained geometric details.

\begin{figure}[t]
    \centering
    \includegraphics[width=0.92\linewidth]{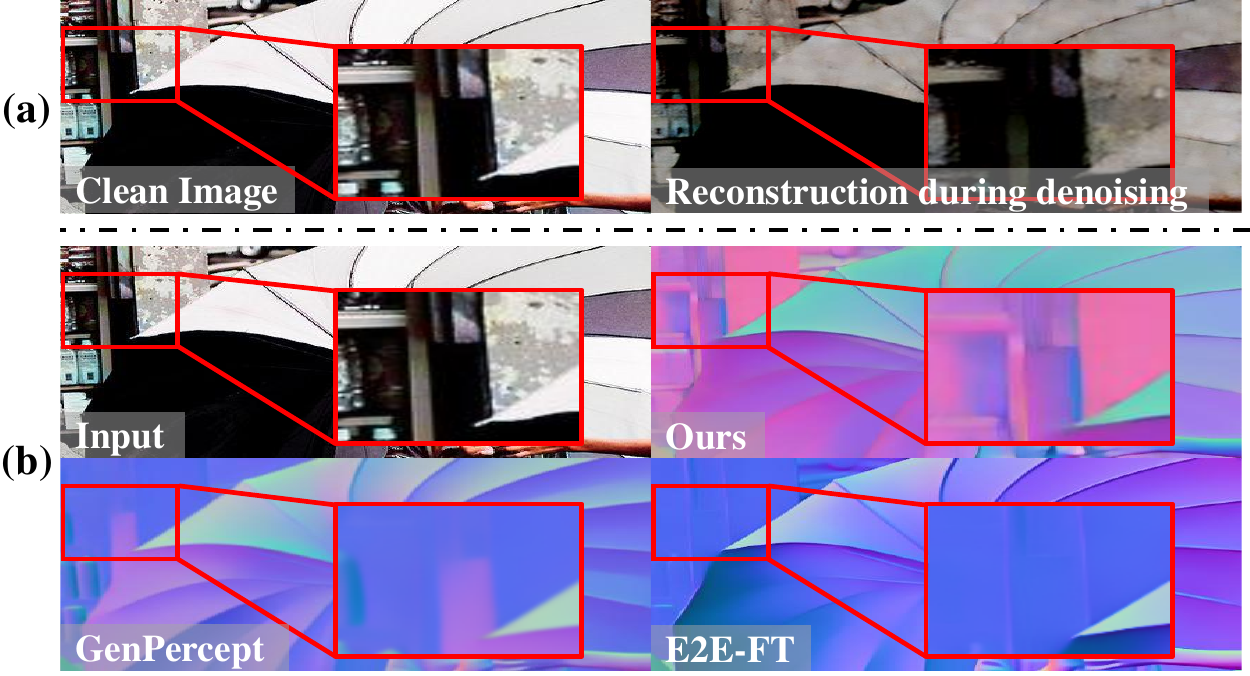}
    \caption{(a) Stochastic noise degarding the integrity of input information; (b) Current attempts to alleviate the stochasticity lead to a loss of geometric fidelity.}
    \label{fig:related}
    \vspace{-1.5em}
\end{figure}

While these diffusion-based dense prediction methods have demonstrated promising results, they still suffer from the stochastic noise inherent in diffusion models. Stochastic noise is a fundamental component of diffusion models, enabling sample diversity that is particularly beneficial for creative image~\cite{li2024photomaker, xu2024diffusion, qin2023unicontrol} and video~\cite{yang2024cogvideox, zhao2024moviedreamer, wang2024framer} generation. However, dense prediction is intrinsically deterministic, indicating that the \textbf{stochastic noise essential to diffusion models may be misaligned with the task's deterministic objective}. We posit that the stochastic noise introduces two critical issues for dense prediction tasks:
1) Stochastic noise disrupts the geometric structures and small-scale objects in the input image, thereby degrading the integrity of input information and impeding precise spatial perception (cf. Fig.~\ref{fig:related} (a));
2) The stochastic noise drives diffusion models to focus on modeling noise distributions~\cite{ho2020denoising} instead of establishing the geometric structure mappings essential for dense prediction tasks~\cite{fu2024geowizard}.
Moreover, the iterative denoising process of the diffusion model further incurs substantial inference overhead.
These observations motivate a fundamental question: \textbf{Can diffusion models be reformulated into a noise-free deterministic framework to better suit dense prediction tasks?}

Recent works attempt to employ deterministic noise to alleviate the stochasticity in diffusion-based dense prediction methods. 
For example, GenPercept~\cite{xu2024matters} and E2E-FT~\cite{garcia2025fine} eliminate stochasticity by fixing the noise schedule (a function of the timestep) to introduce deterministic noise. However, diffusion models learn timestep-specific objectives~\cite{balaji2022ediff, biroli2024dynamical}, leading to distinct diffusion priors at each timestep. Consequently, fixing the noise schedule disrupts this prior structure, resulting in incomplete priors and a loss of geometric fidelity (cf. Fig.~\ref{fig:related} (b)). On the other hand, StableNormal~\cite{ye2024stablenormal} suppresses stochasticity while preserving diffusion priors via a more complex two-stage pipeline with external DINO~\cite{oquab2023dinov2} guidance. However, this design incurs a higher computational overhead.

In this work, we aim to fully eliminate the adverse effects of stochastic noise on diffusion-based dense prediction, without compromising the diffusion prior and with minimal additional computational cost.
To this end, we propose \textbf{D³-Predictor}, a noise-free \underline{d}eterministic \underline{d}iffusion-based \underline{d}ense prediction model initialized from the pretrained diffusion model.
Specifically, we treat the pretrained diffusion models at different timesteps as an ensemble of visual experts following the CleanDIFT~\cite{stracke2025cleandift} paradigm, each exhibiting distinct timestep-dependent diffusion priors.
In this context, each visual expert takes a noisy image together with its timestep as input, while our D³-Predictor operates directly on a clean image.
D³-Predictor then aggregates diffusion priors from multiple visual experts into a complete and noise-free one in a lightweight self-supervised manner, by aligning its internal representations with those of visual experts. 
We simultaneously apply task-specific supervision to the D³-Predictor to easily leverage this aggregated diffusion prior for various dense prediction tasks.

We conduct extensive experiments across multiple dense prediction tasks. The results show that our noise-free D³-Predictor achieves competitive or state-of-the-art performance across diverse zero-shot benchmarks.
More importantly, D³-Predictor markedly improves both training and inference efficiency of diffusion-based dense prediction methods, allowing the model to train with less than half of the data previously required (cf. Fig.~\ref{fig:teaser}) and to efficiently perform inference in a single step.
We believe that D³-Predictor paves a promising path toward reformulating powerful diffusion models for dense prediction.

Our main contributions can be summarized as follows:

\begin{itemize}
    \item  We introduce D³-Predictor, an end-to-end noise-free deterministic diffusion-based dense prediction model. D³-Predictor combines self-supervised noise-free diffusion prior aggregation with task-specific supervision, allowing the model to transfer the diffusion priors into a noise-free geometric prior specifically for dense prediction tasks.
    \item Our D³-Predictor fully eliminates noise in diffusion-based dense prediction, successfully preserving complete diffusion priors with minimal additional computational cost.
    \item Extensive experiments on various dense prediction tasks demonstrate that our noise-free D³-Predictor achieves competitive or state-of-the-art performance.
    It also improves training and inference efficiency while preserving high geometric fidelity of the predicted results.
\end{itemize}

%% file: sec/2_related_work.tex
\section{Related Work}
\label{sec:related}


\begin{figure*}[t]
    \centering
    \includegraphics[width=0.9\linewidth]{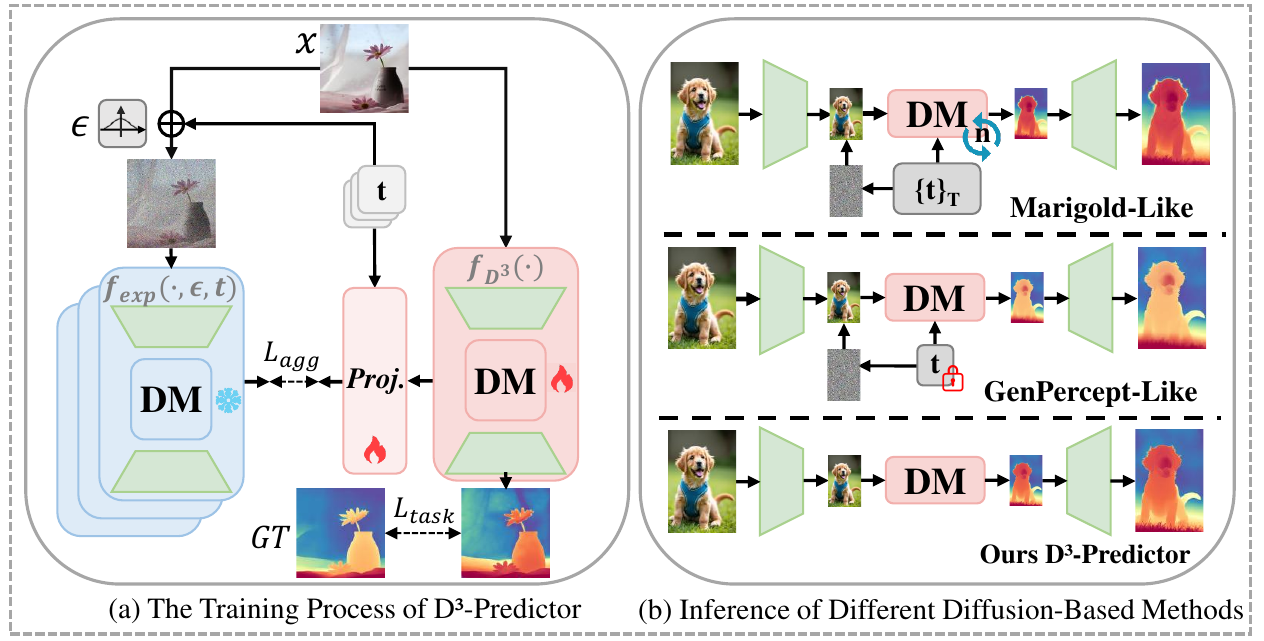}
    \vspace{-0.6em}
    \caption{Overview of the D³-Predictor. (a) We reformulate the pretrained diffusion model into a noise-free framework to better suit dense prediction tasks, without compromising the diffusion prior with minimal overhead. (b) The D³-Predictor takes a clean image as input and produces an accurate prediction with impressive geometric details in a single step.}
    \label{fig:method}
    \vspace{-0.8em}
\end{figure*}

\subsection{Diffusion-based Dense Prediction}
Several works have explored leveraging diffusion models for dense prediction. Methods such as DIFT~\cite{tang2023emergent} and CleanDIFT~\cite{stracke2025cleandift} extract diffusion priors as general-purpose visual features and train lightweight probes for dense prediction tasks, but they are not end-to-end and their performance remains limited. While CleanDIFT~\cite{stracke2025cleandift} obtains noise-free general-purpose features, our method integrates this mechanism with task-specific supervision in an end-to-end manner. This joint optimization allows us to effectively transform the aggregated diffusion priors into noise-free geometric priors specifically tailored for dense prediction. On the other hand, other methods~\cite{ke2024repurposing, gui2025depthfm, zhao2025diception, fu2024geowizard} fine-tune a pretrained diffusion model for dense prediction tasks, achieving impressive results.
However, these fine-tuning methods rely on stochastic noise to exploit diffusion priors. 
GenPercept~\cite{xu2024matters} and E2E-FT~\cite{garcia2025fine} attempt to alleviate stochasticity but disrupt the integrity of diffusion priors, while StableNormal~\cite{ye2024stablenormal} suppresses stochasticity at the cost of additional computational overhead.
In contrast, our method reformulates the diffusion model into a noise-free framework to better suit dense prediction in a simple self-supervised manner, ensuring both prediction performance and speed. 

\subsection{Self-Supervised Enhanced Diffusion Models}
Recent studies~\cite{yu2024representation, wu2025representation, jiang2025no} have shown that enhancing the internal representations of diffusion models, which serve as diffusion priors, can accelerate generative training and improve generation quality. Unlike methods~\cite{yu2024representation} that rely on external pretrained features (e.g., DINO~\cite{oquab2023dinov2}), SRA~\cite{jiang2025no} improves diffusion representations in a self-supervised way via self-representation alignment. The self-supervised enhancement in SRA~\cite{jiang2025no} leads to improvements in both image generation quality and efficiency, demonstrating that diffusion models can provide representation guidance by themselves. In this work, we extend the idea of this self-supervised enhancement from diffusion-based generation to diffusion-based dense prediction. We align the noise-free internal representations of the diffusion model with their noisy counterparts at different noise levels in a self-supervised manner. This process produces complete, noise-free diffusion priors that are inherently more suitable for dense prediction.

%% file: sec/3_method.tex
\section{Method}

\subsection{Preliminary: Diffusion-based Dense Prediction}
In general, a diffusion model consists of an autoencoder and a denoising network. The autoencoder converts an image \( y \in \mathbb{R}^{H \times W \times 3} \) into \( z^{(y)} \in \mathbb{R}^{h \times w \times d} \), where \( (h, w) \) denote the latent spatial dimensions and \( d \) is the latent channel dimension. The diffusion process then gradually corrupts the latent code \( z^{(y)} \) over multiple timesteps by adding noise:
\begin{equation}
z_t^{(y)} = \sqrt{\alpha_t} z^{(y)} + \sqrt{1 - \alpha_t} \epsilon,
\end{equation}
where \(\epsilon \sim \mathcal{N}(0, I)\) is the noise, and \(\alpha_t\) is the noise schedule (a function of the timestep \( t \)).
The denoising network then learns to iteratively reverse this process conditioned on the inputs 
\( c \) (e.g., text prompts), to recover the original latent code from the noisy version.

As illustrated in Fig.~\ref{fig:method} (b), previous diffusion-based dense prediction methods, such as Marigold~\cite{ke2024repurposing}, adopt the time-consuming iterative denoising process described above, where the conditional input \( c \) is an RGB image. 
At each iteration, the model progressively removes stochastic noise associated with a specific noise level and timestep, thus necessitating multi-step inference.

In contrast, recent work GenPercept~\cite{xu2024matters} simplifies this process by fixing the noise schedule. However, it is worth noting that diffusion models have different objectives at different noise levels~\cite{balaji2022ediff, biroli2024dynamical}. At higher noise levels, the model captures coarse structural information, while at lower noise levels, it focuses on details~\cite{rissanen2022generative}. Thus, the model learns distinct diffusion priors across timesteps, which together constitute a complete diffusion prior. Fixing the noise schedule disrupts this prior structure, leading to incomplete priors and suboptimal performance.

Based on these observations, our method aggregates the diffusion priors across timesteps into a noise-free and complete one, which not only preserves the integrity of the diffusion prior but also achieves superior performance through efficient single-step inference.

\subsection{D³-Predictor}
\label{sec:D³-Predictor}
We aim to reformulate diffusion models into a noise-free framework when applied to dense prediction, without compromising diffusion priors with minimal overhead. To this end, we propose D³-Predictor, a noise-free deterministic diffusion-based dense prediction model, as shown in Fig.~\ref{fig:method}.

\subsubsection{Self-Supervised Diffusion Priors Aggregation}
The core of our approach is to train our D³-Predictor $f_{D^3}(\cdot)$ to incorporate a noise-free and complete diffusion prior compared to the pre-trained diffusion model, which is enhanced to be more suitable for deterministic dense prediction tasks. We achieve this enhancement through a simple self-supervised manner, and crucially, relying solely on the model itself rather than costly external guidance (e.g., DINO~\cite{oquab2023dinov2}) adapted by previous works~\cite{ye2024stablenormal}. We begin by initializing D³-Predictor with a pretrained diffusion model.

As shown in Fig~\ref{fig:method} (a), the insight here is to view the diffusion model as an ensemble of \( t \) visual experts, where each expert \( f_{\text{exp}}(\cdot, \epsilon, t) \) exhibits a distinct timestep-dependent diffusion prior. We inherit the mechanism of aligning a clean backbone with a frozen diffusion model from CleanDIFT~\cite{stracke2025cleandift}. 
While the visual experts take a noisy image \( (\epsilon, x) \) together with its timestep \( t \) as input, our D³-Predictor operates directly on a clean image \( x \). During training, D³-Predictor aligns its internal representations \( r_{D^3}^{K} \) = \( f_{D^3}^{K}(x) \) with those of \( t \) visual experts \( r_{\text{exp}}^{K, t} \) = \( f_{\text{exp}}^{K}(x, \epsilon, t) \). Here, \( K \) is the index set of model layers selected for alignment.

To stabilize this timestep-dependent representation alignment, we introduce a lightweight, trainable timestep-conditioned projection head $P_{\varphi}^{K} (\cdot, t)$ parameterized by $\varphi$. While the use of projection heads for representation alignment has proven effective in prior works~\cite{yu2024representation, wu2025representation, jiang2025no, stracke2025cleandift}, we specifically leverage the timestep-conditioned projection head adopted by CleanDIFT~\cite{stracke2025cleandift} and shown in Fig~\ref{fig:method} (a).
This projection head serves to project the single, noise-free representation $r_{D^3}^{K}$ into $T$ distinct timestep-dependent representation spaces for the purpose of matching the expert representations. This mechanism implicitly requires our model's representation to encode the diverse diffusion priors corresponding to different noise levels. We measure the alignment by their point-wise distance:
\begin{equation}
\label{dist}
L_{\text{agg}} =
\sum_{t \in T} \sum_{k \in K}
\text{dist}\!\left(
r_{\text{exp}}^{k, t},
P_{\varphi}^{k}(\mathbf{r}_{\text{D}^{3}}^{k}, t)
\right),
\end{equation}
Where \( \text{dist}(\cdot, \cdot) \) denotes a predefined distance function. By minimizing \( L_{\text{agg}} \), our D³-Predictor obtains a noise-free and complete internal representation that aggregates diffusion model's timestep-dependent priors into a single one, which is more suitable for deterministic dense prediction tasks.

\subsubsection{Task-Specific Supervision}
We meanwhile apply task-specific supervision by minimizing the task-specific loss \( L_{\text{task}} \) to the D³-Predictor to leverage this aggregated diffusion prior for dense prediction tasks. For depth estimation, \( L_{\text{task}} \) is defined as a weighted sum of the MSE loss, the affine-invariant loss~\cite{ranftl2020towards}, and the gradient loss in log space~\cite{ranftl2020towards}. For surface normal estimation, \( L_{\text{task}} \) is defined as a weighted sum of the MSE loss and the angular loss. For image matting, \( L_{\text{task}} \) is defined as a weighted sum of the MSE loss, \( \ell_1 \) loss the and gradient loss. Notably, the task-specific supervision and the self-supervised diffusion priors aggregation are jointly optimized. As a result, the aggregated diffusion prior preserves more task-relevant geometric structure information while suppressing irrelevant color-related information. Finally, the overall training objective \( L \) is defined as:
\begin{equation}
L = L_{\text{agg}} + \lambda L_{\text{task}},
\end{equation}
where \( \lambda > 0 \) is a hyperparameter that balances the self-supervised and supervised training objectives.

During inference, the D³-Predictor takes a clean image as the sole input and generates an accurate prediction result with impressive geometric details in a single step, as illustrated in Fig.~\ref{fig:method} (b). More details (e.g., calculation details of \( L_{\text{task}} \) for different tasks) can be found in the Appendix.

\subsection{Training on Mixed Data}
The quality of real dense prediction datasets suffer from the physical limitations of sensing devices~\cite{huang2023neural}.
Recent dense prediction methods~\cite{ke2024repurposing, zhao2025diception, xu2024matters} have instead opted to train entirely on synthetic datasets.
However, we argue that the domain gap between synthetic and real data can hinder real-world generalization.
In this work, we train our model on a balanced mixture of synthetic data and real images with pseudo labels.
The synthetic data provide fine-grained geometric details, whereas the real images contribute to the model’s generalization by providing more diverse scenes.

%% file: sec/4_experiment.tex
\begin{table*}[t]
    \centering
    \scriptsize
    \setlength{\tabcolsep}{3pt}
    \renewcommand{\arraystretch}{1.2}
    \caption{Quantitative comparison of different depth estimation methods across multiple zero-shot benchmarks. We use the same evaluation protocol as DICEPTION~\cite{zhao2025diception}. NFEs is the numbers of function evaluations (ensemble $\times$ steps)}
    \vspace{-0.6em}
    \label{tab:depth_comparison}
    \begin{adjustbox}{width=\textwidth}
    \begin{tabular}{lccccccccccccc}
    \toprule
    \multirow{2}{*}{Method} & \multirow{2}{*}{\makecell{Training\\ Samples}} & \multirow{2}{*}{\makecell{NFEs}} &
    \multicolumn{2}{c}{KITTI} & \multicolumn{2}{c}{NYU} & \multicolumn{2}{c}{ScanNet} &
    \multicolumn{2}{c}{DIODE} & \multicolumn{2}{c}{ETH3D} & \multirow{2}{*}{\makecell{Average\\ Rank$\downarrow$}} \\
    \cmidrule(lr){4-5}\cmidrule(lr){6-7}\cmidrule(lr){8-9}\cmidrule(lr){10-11}\cmidrule(lr){12-13}
    & & & AbsRel$\downarrow$ & $\delta_1\uparrow$ & AbsRel$\downarrow$ & $\delta_1\uparrow$ & AbsRel$\downarrow$ & $\delta_1\uparrow$ & AbsRel$\downarrow$ & $\delta_1\uparrow$ & AbsRel$\downarrow$ & $\delta_1\uparrow$ & \\
    \midrule
    MiDaS~\cite{ranftl2020towards} & 2M & - & 0.236 & 0.630 & 0.111 & 0.885 & 0.121 & 0.846 & 0.332 & 0.715 & 0.184 & 0.752 & 6.9 \\
    Omnidata~\cite{eftekhar2021omnidata} & 12.2M & - & 0.149 & 0.835 & 0.074 & 0.945 & 0.075 & 0.936 & 0.339 & 0.742 & 0.166 & 0.778 & 5.1 \\
    DPT-large~\cite{ranftl2021vision} & 1.4M & - & 0.100 & 0.901 & 0.098 & 0.903 & 0.082 & 0.934 & 0.182 & 0.758 & 0.078 & 0.946 & 4.2 \\
    DiverseDepth~\cite{yin2020diversedepth} & 320K & - & 0.190 & 0.704 & 0.117 & 0.875 & 0.109 & 0.882 & 0.376 & 0.631 & 0.228 & 0.694 & 7.0 \\
    HDN~\cite{zhang2022hierarchical} & 300K & - & 0.115 & 0.867 & 0.069 & 0.948 & 0.080 & 0.939 & 0.246 & 0.780 & 0.121 & 0.833 & 3.9 \\
    DepthAnything~\cite{yang2024depth} & 63.5K & - & 0.080 & 0.946 & 0.043 & \textbf{0.980} & 0.043 & 0.981 & 0.261 & 0.759 & 0.058 & \textbf{0.984} & 2.2 \\
    DepthAnything v2~\cite{yang2024depthv2} & 62.6M & - & 0.080 & 0.943 & 0.043 & 0.979 & 0.042 & 0.979 & 0.321 & 0.758 & 0.066 & 0.983 & 2.8 \\
    Metric3D v2~\cite{hu2024metric3d} & 16M & - & \textbf{0.052} & \textbf{0.979} & \textbf{0.039} & 0.979 & \textbf{0.023} & \textbf{0.989} & \textbf{0.147} & \textbf{0.892} & \textbf{0.040} & 0.983 & \textbf{1.2} \\
    \midrule
    Unified-IO~\cite{lu2022unified} & 48K & - & 0.188 & 0.699 & \textbf{0.059} & \textbf{0.970} & \textbf{0.063} & \textbf{0.965} & 0.369 & \textbf{0.906} & 0.103 & 0.906 & 1.9 \\
    4M-XL~\cite{mizrahi20234m} & 759M & - & 0.105 & 0.896 & 0.068 & 0.951 & 0.065 & 0.955 & \textbf{0.331} & 0.734 & \textbf{0.070} & \textbf{0.953} & \textbf{1.7} \\
    OneDiffusion~\cite{le2025one} & 500K & - & \textbf{0.101} & \textbf{0.908} & 0.087 & 0.924 & 0.094 & 0.906 & 0.399 & 0.661 & 0.072 & 0.949 & 2.4 \\
    \midrule
    Marigold~\cite{ke2024repurposing} & 74K & $10\times50$ & 0.099 & 0.916 & 0.055 & 0.964 & 0.064 & 0.951 & 0.308 & 0.773 & 0.065 & 0.960 & 4.4 \\
    DMP~\cite{lee2024exploiting} & 10K & $1\times5$ & 0.240 & 0.622 & 0.109 & 0.891 & 0.146 & 0.814 & 0.361 & 0.706 & 0.128 & 0.857 & 7.7 \\
    DepthFM~\cite{gui2025depthfm} & 63K & $10\times4$ & 0.174 & 0.718 & 0.082 & 0.932 & -- & -- & \textbf{0.225} & \textbf{0.800} & -- & -- & 5.2 \\
    Deception~\cite{zhao2025diception} & 500K & $1\times28$ & \textbf{0.075} & \textbf{0.945} & 0.072 & 0.939 & 0.075 & 0.938 & \underline{0.243} & 0.741 & \textbf{0.053} & \textbf{0.967} & 3.6 \\
    GeoWizard~\cite{fu2024geowizard} & 280K & $10\times50$ & 0.129 & 0.851 & 0.059 & 0.959 & 0.066 & 0.953 & 0.328 & 0.753 & 0.077 & 0.940 & 5.6 \\
    Lotus-G~\cite{he2024lotus} & 59K & $1\times1$ & 0.113 & 0.877 & \underline{0.054} & \underline{0.966} & 0.060 & 0.960 & -- & -- & \underline{0.062} & \underline{0.961} & 3.3 \\
    E2E-FT~\cite{garcia2025fine} & 74K & $1\times1$ & 0.096 & 0.921 & \underline{0.054} & 0.965 & \underline{0.058} & \textbf{0.965} & 0.303 & 0.776 & 0.064 & 0.959 & \underline{3.1} \\
    Genpercept~\cite{xu2024matters} & 90K & $1\times1$ & 0.094 & 0.923 & 0.091 & 0.932 & \textbf{0.056} & \textbf{0.965} & 0.302 & 0.767 & 0.066 & 0.957 & 4.0 \\
    \rowcolor{gray!14}
    Ours & 30K & $1\times1$ & \underline{0.082} & \underbar{0.940} & \textbf{0.052} & \textbf{0.970} & 0.063 & \underline{0.963} & 0.290 & \underline{   0.779} & \underline{0.062} & \underline{0.961} & \textbf{2.1} \\
    \bottomrule
    \end{tabular}
    \end{adjustbox}
\end{table*}

\begin{table*}[t]
    \centering
    \scriptsize
    \setlength{\tabcolsep}{3pt}
    \renewcommand{\arraystretch}{1.2}
    \caption{Quantitative comparison of different surface normal estimation methods across multiple zero-shot benchmarks. We use the same evaluation protocol as Marigold-Normals~\cite{ke2025marigold}. NFEs is the numbers of function evaluations (ensemble $\times$ steps).}
    \vspace{-0.6em}
    \label{tab:normal_comparison}
    \begin{adjustbox}{width=\textwidth}
    \begin{tabular}{lcccccccccccccc}
    \toprule
    \multirow{2}{*}{Method} & \multirow{2}{*}{\makecell{Training\\ Samples}} & \multirow{2}{*}{\makecell{NFEs}} &
    \multicolumn{2}{c}{NYUv2} & \multicolumn{2}{c}{ScanNet} & \multicolumn{2}{c}{iBims-1} &
    \multicolumn{2}{c}{DIODE} & \multicolumn{2}{c}{OASIS} & \multirow{2}{*}{\makecell{Average\\ Rank$\downarrow$}} \\
    \cmidrule(lr){4-5}\cmidrule(lr){6-7}\cmidrule(lr){8-9}\cmidrule(lr){10-11}\cmidrule(lr){12-13}
    & & & Mean$\downarrow$ & 11.25$^\circ$$\uparrow$ & Mean$\downarrow$ & 11.25$^\circ$$\uparrow$ &
    Mean$\downarrow$ & 11.25$^\circ$$\uparrow$ & Mean$\downarrow$ & 11.25$^\circ$$\uparrow$ &
    Mean$\downarrow$ & 11.25$^\circ$$\uparrow$ & \\
    \midrule
    DSINE~\cite{bae2024rethinking} & 160K & - & 0.164 & 0.596 & \textbf{0.162} & \textbf{0.610} & \textbf{0.171} & 0.674 & 0.199 & 0.418 & 0.244 & \textbf{0.288} & 1.7 \\
    Omnidata v2~\cite{kar20223d} & 12.2M & - & 0.172 & 0.555 & \textbf{0.162} & 0.602 & 0.182 & 0.639 & 0.206 & 0.408 & 0.242 & 0.277 & 2.5 \\
    Metric3D v2~\cite{hu2024metric3d} & 16M & - & \textbf{0.133} & \textbf{0.664} & -- & -- & 0.196 & \textbf{0.697} & \textbf{0.126} & \textbf{0.649} & \textbf{0.234} & 0.285 & \textbf{1.4} \\
    \midrule
    GeoWizard~\cite{fu2024geowizard} & 280K & $10\times50$ & 0.190 & 0.500 & 0.176 & 0.546 & 0.193 & 0.623 & 0.247 & 0.301 & 0.253 & 0.269 & 6.0 \\
    StableNormal~\cite{ye2024stablenormal} & 280K & $1\times10$ & 0.178 & 0.542 & 0.167 & 0.540 & \underline{0.171} & 0.678 & \underline{0.193} & \textbf{0.538} & 0.257 & 0.254 & 4.1 \\
    Marigold-Normals~\cite{ke2025marigold} & 77K & $1\times1$ & \underline{0.164} & 0.589 & \underline{0.149} & \underline{0.643} & 0.172 & 0.656 & 0.195 & 0.432 & \textbf{0.232} & \textbf{0.283} & 2.7 \\
    Lotus-G~\cite{he2024lotus} & 59K & $1\times1$ & 0.169 & 0.591 & 0.153 & \underline{0.640} & 0.174 & 0.661 & 0.212 & 0.397 & 0.247 & 0.270 & 3.7 \\
    E2E-FT~\cite{garcia2025fine} & 74K & $1\times1$ & 0.165 & \textbf{0.604} & \textbf{0.147} & \textbf{0.661} & \textbf{0.161} & \textbf{0.697} & \textbf{0.190} & 0.444 & \underline{0.236} & \underline{0.279} & \textbf{1.6} \\
    Genpercept~\cite{xu2024matters} & 44K & $1\times1$ & 0.183 & 0.560 & 0.182 & 0.574 & 0.183 & 0.638 & 0.223 & 0.381 & 0.259 & 0.233 & 5.7 \\
    \rowcolor{gray!15}
    Ours & 30K & $1\times1$ & \textbf{0.162} & \underline{0.595} & 0.153 & 0.618 & \textbf{0.161} & \underline{0.687} & \textbf{0.190} & \underline{0.472} & \textbf{0.232} & 0.276 & \underline{2.0} \\
    \bottomrule
    \end{tabular}
    \end{adjustbox}
\end{table*}

\begin{figure*}[t]
    \centering
    \includegraphics[width=0.9\linewidth]{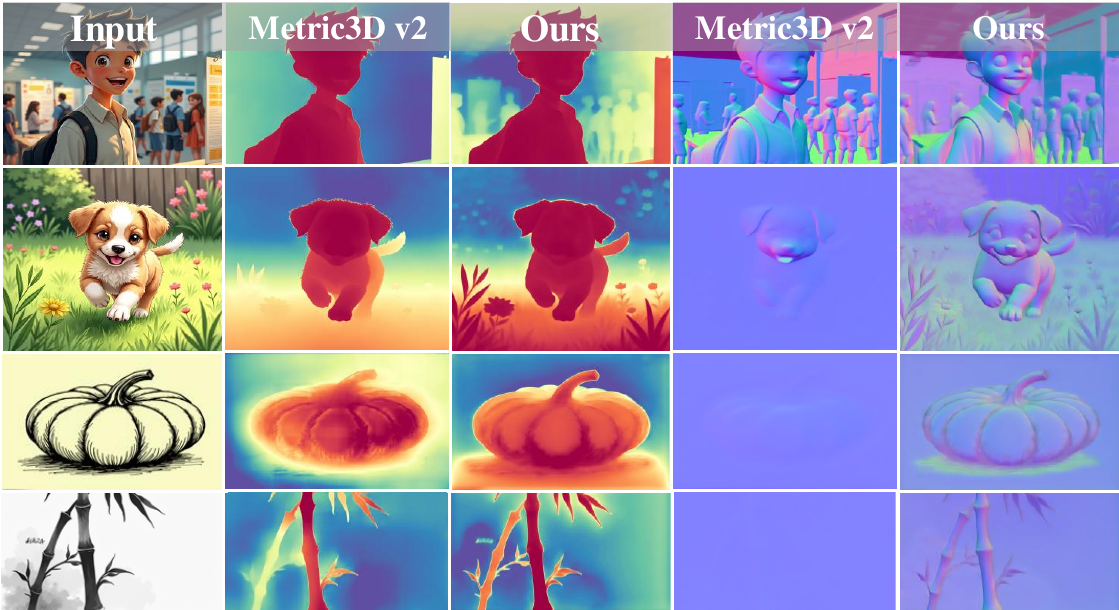}
    \caption{Comparison of geometric detail generation capability on out-of-domain scenes. State-of-the-art discriminative models show significant performance degradation on out-of-domain images, while our model works surprisingly well.}
    \vspace{-0.6em}
    \label{fig:experiment}
\end{figure*}

\section{Experiments}
\subsection{Experimental Settings}
\subsubsection{Training Datasets}
\label{sec:Training Datasets}
We train D³-Predictor on different datasets for different dense prediction tasks: 1) For depth estimation: we use $10,777$ samples from Hypersim~\cite{roberts2021hypersim}, a synthetic indoor scene dataset; $4,252$ samples from Virtual KITTI~\cite{cabon2020virtual}, a synthetic driving scene dataset; and $15,160$ real images from COCO~\cite{lin2014microsoft} with pseudo-depth labels generated by Depth Pro~\cite{bochkovskii2024depth}. 2) For surface normal estimation: we use $29,284$ samples from Hypersim~\cite{roberts2021hypersim} and $376$ samples from Sintel~\cite{butler2012naturalistic}, which contains both indoor and outdoor scenes from a short animated film. 3) For image matting: we use the training set of P3M-10K~\cite{li2021privacy}, which includes $9,421$ images and is currently the largest portrait matting dataset.
We only employ real images with pseudo labels for depth estimation, since other tasks have been less explored and their models generalize poorly to in-the-wild images. More details on the training data (e.g. data mixing strategies) are in Appendix.
\subsubsection{Implementation Details}
Unless otherwise specified, we initialize our D³-Predictor with Stable Diffusion v2.1~\cite{rombach2022high} for fair comparison with previous works that also adopt it.
We fine-tune the U-Net of it using an image resolution of $768\times768$.
We employ the Adam optimizer~\cite{loshchilov2017decoupled} with a batch size of 32 and a learning rate of $1e-5$, following a linear warm-up schedule. All experiments are conducted on two NVIDIA L40S GPUs.

\subsection{Comparison to the state-of-the-art}
\subsubsection{Depth Estimation}
Depth estimation predicts the distance between the object and the camera from an image.
We perform evaluations on five zero-shot real-world benchmarks, including NYUv2~\cite{silberman2012indoor}, KITTI~\cite{geiger2013vision}, ETH3D~\cite{schops2017multi}, ScanNet~\cite{dai2017scannet}, and DIODE~\cite{vasiljevic1908diode}. 
The evaluation metrics include the Absolute Mean Relative Error (RelAbs) and the $\delta_1$-accuracy. 

We compare specialized discriminative models~\cite{hu2024metric3d}, multi-task models~\cite{le2025one}, and diffusion-based models~\cite{ke2024repurposing}. Table~\ref{tab:depth_comparison} presents the quantitative comparison results.
D³-Predictor demonstrates remarkable zero-shot generalization and promising performance across five benchmarks.
Compared with diffusion-based methods, D³-Predictor requires neither multi-step sampling nor test-time ensembling, thereby achieving significantly higher inference efficiency and stability without sacrificing accuracy.
Although our model lags behind state-of-the-art discriminative models in terms of quantitative performance, it produces visually sharper results, as shown in Fig.~\ref{fig:experiment}.
When encountering out-of-domain scenes (e.g., AIGC-generated images), the performance of discriminative models drops drastically.
Notably, most discriminative models require large-scale datasets and complex data pipelines, while our model only relies on limited synthetic data and arbitrary real images.

\subsubsection{Surface Normal Estimation}
Surface normal estimation predicts the direction vector perpendicular to the surface at each point on the object. We perform evaluations on five zero-shot real-world benchmarks, including NYUv2~\cite{silberman2012indoor}, ScanNet~\cite{dai2017scannet}, iBims-1~\cite{koch2018evaluation}, DIODE~\cite{vasiljevic1908diode} and OASIS~\cite{chen2020oasis}. The evaluation metrics include mean angular error and the percentage of pixels with an angular error below the threshold of $11.25^\circ$.

We compare specialized discriminative models~\cite{hu2024metric3d} and diffusion-based models~\cite{ke2025marigold}. The quantitative results in Table~\ref{tab:normal_comparison} reveal that our D³-Predictor is competitive with state-of-the-art diffusion-based models across five benchmarks, while also demonstrating impressive inference efficiency and stability. Although E2E-FT~\cite{garcia2025fine} achieves marginally better performance than D³-Predictor, our model is significantly more data-efficient, requiring less than half of its training data. This is particularly advantageous in real-world applications, where annotating large-scale dense labels and training diffusion models is costly and thus often limits deployment. 
Fig.~\ref{fig:experiment} further highlights the geometric detail generation capability and robustness of our approach.

\begin{figure*}[t]
    \centering
    \includegraphics[width=0.9\linewidth]{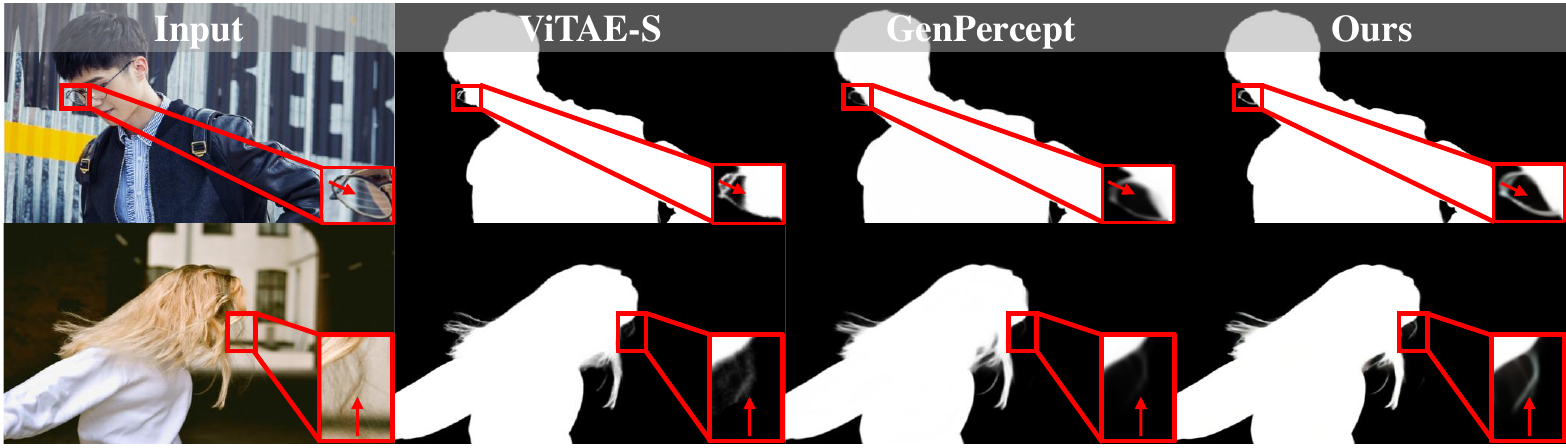}
    \caption{Qualitative comparison of image matting on the P3M-500-NP test set.}
    \label{fig:experiment_matting}
    \vspace{-1.0em}
\end{figure*}

\subsubsection{Image Matting}
Image matting aims to extract the foreground and background from an image and generate an alpha matte. We evaluate our method on the P3M-500-NP test set~\cite{li2021privacy}. The evaluation metrics include the sum of absolute differences (SAD), mean squared error (MSE), mean absolute difference (MAD) and connectivity (Conn.).

Table~\ref{tab:results_matte} presents the quantitative comparison results. Overall, our method significantly outperforms a recent diffusion-based method GenPercept~\cite{xu2024matters}, demonstrating the superiority of our D³-Predictor for image matting. Moreover, compared with state-of-the-art discriminative matting models, D³-Predictor achieves clearly competitive performance. Qualitative comparisons in Fig.~\ref{fig:experiment_matting} further show that our method produces high-quality alpha mattes, especially in challenging regions such as transparent area and hair.

\begin{table}[htbp]
    \vspace{-0.2em}
    \centering
    \scriptsize
    \setlength{\tabcolsep}{6pt}
    \renewcommand{\arraystretch}{1.2}
    \caption{Evaluation of image matting on P3M-500-NP test set.}
    \vspace{-0.8em}
    \label{tab:results_matte}
    \resizebox{0.95\columnwidth}{!}{%
      \begin{tabular}{lcccccc}
      \toprule
      Method & \makecell{Training\\ Samples} & SAD$\downarrow$ & MAD$\downarrow$ & MSE$\downarrow$ & CONN$\downarrow$ & \makecell{Average\\ Rank$\downarrow$}\\
      \midrule
      HATT~\cite{qiao2020attention}&    60K      & 30.35 & 0.0176 & 0.0072 & 27.42 & 4.8 \\
      SHM~\cite{chen2018semantic}&     53K     & 20.77 & 0.0122 & 0.0093 & 17.09 & 4.3 \\
      MODNet~\cite{ke2022modnet}&   45K  & 16.70 & 0.0097 & 0.0051 & 13.81 & 3 \\
      P3M-Net~\cite{li2021privacy}&   9K   & 11.23 & 0.0065 & 0.0035 & 12.51 & 2 \\
      ViTAE-S~\cite{ma2023rethinking}&  9K  & \textbf{7.59} & \textbf{0.0044} & \textbf{0.0019} & \textbf{6.96} & \textbf{1} \\
      \midrule
      GenPercept~\cite{xu2024matters}& 9K  & 12.77 & 0.0074 & 0.0027 & 10.46 & 2 \\
      \rowcolor{gray!15}
      Ours& 9K  & \textbf{7.97} & \textbf{0.0046} & \textbf{0.0015} & \textbf{7.43} & \textbf{1} \\
      \bottomrule
      \end{tabular}%
    }
    \vspace{-1.0em}
\end{table}

\subsubsection{Generalization to Domain-Specific Tasks}
\label{sec:Generalization to Domain-Specific Tasks}
To further assess the generalization and practical utility of our D³-Predictor, we evaluate it on 20 additional domain-specific dense prediction tasks, such as medical image analysis. These tasks are typically constrained by data scarcity, often due to factors like privacy protection. For these evaluations, we train our model using only $15$ samples per task with a simple MSE loss. We present \textit{Spinal Morphology Assessment} source from \cite{chu2015annotated} as a representative example.

\begin{figure}[t]
    \vspace{-0.3em}
    \centering
    \includegraphics[width=0.7\linewidth]{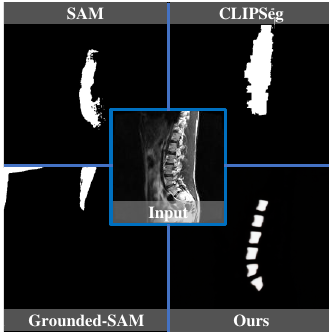}
    \caption{Comparison on the \textit{Spinal Morphology Assessment} task.}
    \vspace{-1.6em}
    \label{fig:spinal_visualization}
\end{figure}

As summarized in Table~\ref{tab:spinal_morphology}, D³-Predictor significantly outperforms strong zero-shot baselines, such as SAM~\cite{kirillov2023segment}, by over $\mathbf{60\%}$ in average performance, demonstrating its superior generalization in the medical domain. The qualitative results in Fig.~\ref{fig:spinal_visualization} further show our model accurately captures the complex structure of the spine, underscoring its excellent fine-grained geometric perception. More comprehensive results on the remaining tasks are in the Appendix.

\begin{table}[htbp]
    \centering
    \scriptsize
    \setlength{\tabcolsep}{6pt}
    \renewcommand{\arraystretch}{1.2}
    \caption{Evaluation of the \textit{Spinal Morphology Assessment} task.}
    \vspace{-0.8em}
    \label{tab:spinal_morphology}
    \begin{tabular}{lccccc}
    \toprule
    Model & \makecell{Training\\ Samples} & IoU $\uparrow$ & PA $\uparrow$ & DiCE $\uparrow$ & Avg. $\uparrow$ \\
    \midrule
    SAM~\cite{kirillov2023segment} & 1.1B & \underline{0.555} & \underline{0.611} & \underline{0.612} & 0.593 \\
    CLIPSeg~\cite{luddecke2022image} & 345K & 0.519 & 0.569 & 0.551 & 0.546 \\
    Grounded-SAM~\cite{ren2024grounded} & 1.1B & 0.464 & 0.481 & 0.481 & 0.475 \\
    \rowcolor{gray!10}
    Ours & 15 & \textbf{0.932} & \textbf{0.969} & \textbf{0.963} & \textbf{0.955} \\
    \bottomrule
    \end{tabular}
\end{table}

\begin{figure*}[t]
    \centering
    \includegraphics[width=0.9\linewidth]{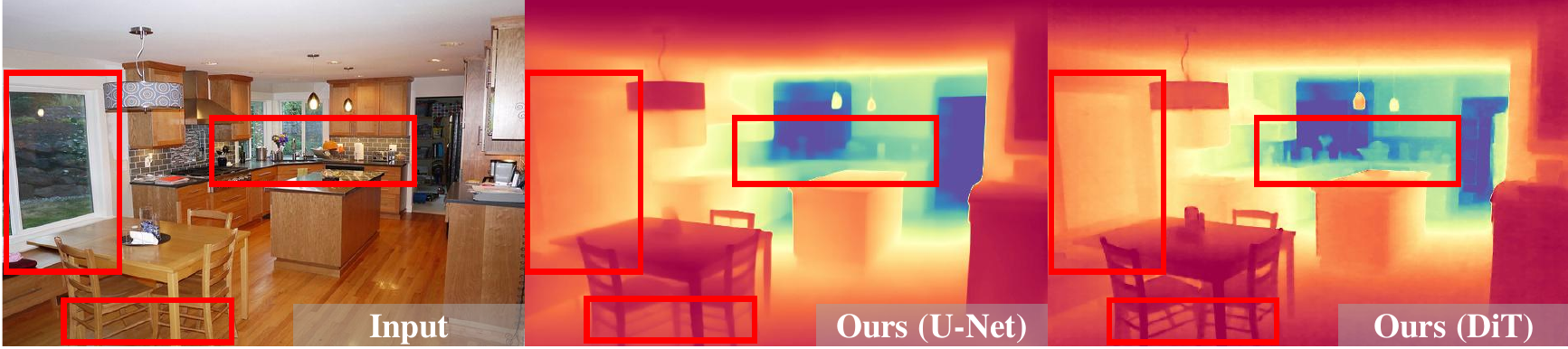}
    \caption{DiT-based D³-Predictor exhibits even higher fidelity, indicating that DiT-based architectures hold great potential for further improving diffusion-based dense prediction performance as more computational resources become available.}
    \label{fig:flux v.s. U-Net}
    \vspace{-0.8em}
\end{figure*}

\subsection{Ablation Study}
\label{sec:Ablation Study}
For simplicity, we conduct ablation studies on the representative task of depth estimation. We report results on two standard benchmarks: NYUv2~\cite{silberman2012indoor} for indoor scenes and KITTI~\cite{geiger2013vision} for driving scenes.
\subsubsection{Training Data Volume}
\begin{figure}[h]
    \vspace{-0.4em}
    \centering
    \includegraphics[width=0.9\linewidth]{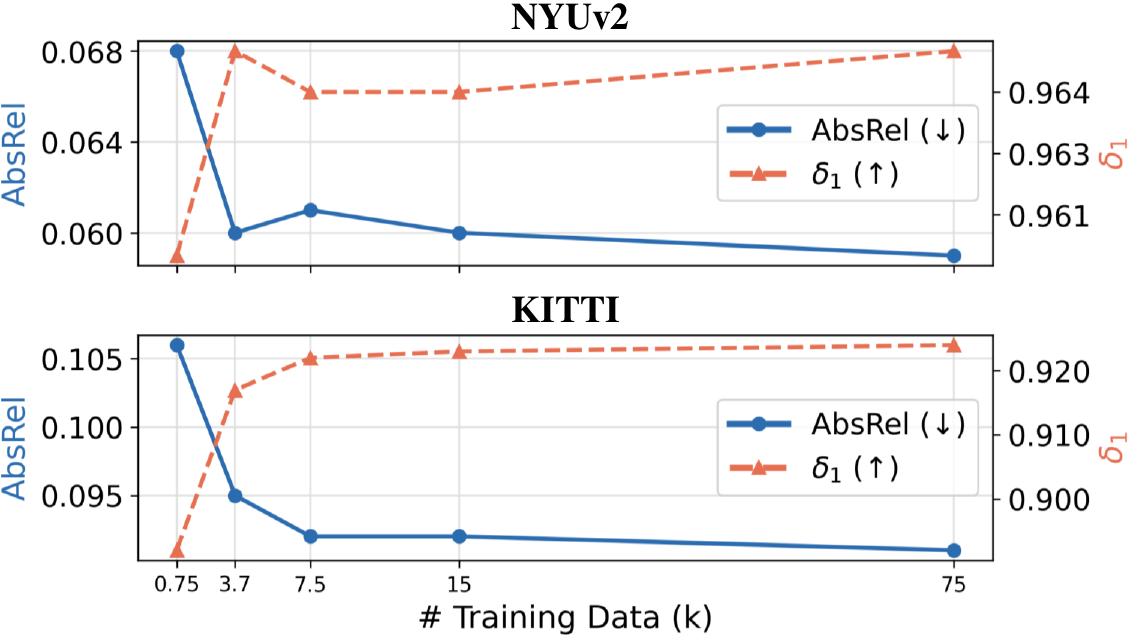}
    \vspace{-0.6em}
    \caption{The ablation study on the training data volume.}
    \label{fig:training_data}
    \vspace{-1.0em}
\end{figure}

Our method achieves a remarkable balance between training data volume and performance. To further evaluate training data efficiency, we train D³-Predictor on varying amounts of synthetic data. As shown in Fig.~\ref{fig:training_data}, D³-Predictor already delivers acceptable performance even with a limited amount of training data, demonstrating the training data efficiency of our approach.
It is worth noting that compared with state-of-the-art discriminative depth estimation models, such as Depth Pro~\cite{bochkovskii2024depth}, our method exhibits superior data efficiency.
First, Depth Pro is trained using a combination of more than twenty heterogeneous datasets, whereas our model only requires a small subset from three datasets.
Second, Depth Pro relies on complex data augmentation pipelines to utilize large-scale mixed data, while our method does not require any specialized data processing.

\subsubsection{Improving Generalization with Real Images}
\label{sec:Improving Generalization with Real Images}
As illustrated in Fig.~\ref{fig:training_data}, further increasing the amount of synthetic training data to 75K yields only marginal performance gains, suggesting that the limited diversity of synthetic data constrains model generalization.
To address this limitation, we train our model on a balanced mixture of synthetic data and real images with pseudo-labels to improve its generalization. This approach is feasible for two main reasons.
First, state-of-the-art discriminative models, although less capable of producing fine-grained visual details than diffusion-based methods, typically achieve superior quantitative performance and can generate reliable pseudo-labels for real images.
Second, diffusion priors, including our noise-free aggregated diffusion prior, inherently originate from Internet-scale datasets~\cite{schuhmann2022laion}, which enables our model to effectively process both synthetic and real data in a unified manner.
Table~\ref{tab:kitti-nyuv2} demonstrates that incorporating real images significantly improves the model’s generalization ability across zero-shot real-world benchmarks.

\begin{table}[h]
    \centering
    \caption{Performance comparison of D³-Predictor under different training data configurations.}
    \label{tab:kitti-nyuv2}
    \resizebox{0.95\columnwidth}{!}{%
      \begin{tabular}{M{3.0cm}*{4}{M{1.4cm}}}
        \toprule
        \multirow{2}{*}{Synthetic + Real} & \multicolumn{2}{c}{KITTI} & \multicolumn{2}{c}{NYUv2} \\
        \cmidrule(lr){2-3}\cmidrule(lr){4-5}
         & AbsRel$\downarrow$ & $\delta_1\uparrow$ & AbsRel$\downarrow$ & $\delta_1\uparrow$ \\
        \midrule
        7.5K + 7.5K & 0.087 & 0.933 & 0.060 & 0.967 \\
        15K + 0     & 0.092 & 0.923 & 0.061 & 0.963 \\
        7.5K + 0    & 0.092 & 0.922 & 0.061 & 0.964 \\
        \bottomrule
      \end{tabular}%
    }
\end{table}



\subsubsection{Architecture of Diffusion Model}

Recent progress in image generation has shifted from the U-Net architecture to the DiT architecture~\cite{peebles2023scalable}, achieving better image quality and fidelity~\cite{flux2024, labs2025flux1kontextflowmatching, wu2025qwen}.
We also initialize our D³-Predictor with FLUX.1 dev~\cite{flux2024}, a DiT-based pretrained diffusion model. The detailed design of the DiT-based D³-Predictor is provided in the Appendix.


Due to the significant increase in model size (over $15\times$) and our limited computational resources, we perform only a quick validation on 7.5K synthetic samples using an input resolution of 768$\times$768. Note that this resolution is sub-optimal for FLUX.1 dev~\cite{flux2024}, which is pretrained on much higher resolutions.  As shown in Table~\ref{tab:architecture}, the DiT-based D³-Predictor achieves competitive quantitative performance compared to the U-Net variant. Furthermore, Fig.~\ref{fig:flux v.s. U-Net} reveals that it exhibits higher perceptual fidelity. These findings suggest that the DiT-based D³-Predictor holds immense potential for advancing dense prediction performance.



\begin{table}[h]
    \centering
    \caption{Performance comparison of D³-Predictor under different diffusion model architectures.}
    \label{tab:architecture}
    \resizebox{0.95\columnwidth}{!}{%
      \begin{tabular}{M{3.0cm}*{4}{M{1.4cm}}}
        \toprule
        \multirow{2}{*}{Architecture} & \multicolumn{2}{c}{KITTI} & \multicolumn{2}{c}{NYUv2} \\
        \cmidrule(lr){2-3}\cmidrule(lr){4-5}
         & AbsRel$\downarrow$ & $\delta_1\uparrow$ & AbsRel$\downarrow$ & $\delta_1\uparrow$ \\
        \midrule
        D³-Predictor (DiT) & 0.095 & 0.918 & 0.065 & 0.959 \\
        D³-Predictor (U-Net)     & 0.092 & 0.922 & 0.061 & 0.964 \\
        \bottomrule
      \end{tabular}%
    }
\end{table}

%% file: sec/5_conclusion.tex
\section{Conclusion}
In this paper, we show that eliminating the stochastic noise inherent in diffusion models is crucial for improving the performance, generalization, and efficiency of diffusion-based dense prediction.
To this end, we propose D³-Predictor, a noise-free deterministic diffusion-based dense prediction model that aggregates timestep-dependent priors from the pretrained diffusion model into a complete, noise-free prior in a self-supervised manner. We achieve this by aligning the internal representations of the D³-Predictor with those of the pretrained diffusion model. This noise-free aggregated diffusion prior can be effectively leveraged for various dense prediction tasks through simple task-specific supervision. Furthermore, D³-Predictor requires less than half the training data previously used, and performs inference in a single step, significantly improving the training efficiency and inference efficiency of diffusion-based dense prediction. Extensive experiments on various dense prediction tasks demonstrate that D³-Predictor achieves superior performance and generalization across various dense prediction tasks.

%% file: sec/X_suppl.tex
\clearpage
\setcounter{page}{1}
\maketitlesupplementary

This supplementary material provides additional details and more qualitative and quantitative analysis to complement the main paper. The content is organized as follows:
\begin{itemize}[leftmargin=*]
    \item More Implementation Details (Appendix~\ref{sec:More Implementation Details})
    \item Additional Visualization Results (Appendix~\ref{sec:Additional Visualization Results})
    \item Additional Experimental Results (Appendix~\ref{sec:Additional Experimental Results})
\end{itemize}

\vspace{0.6em}


\begin{figure}[b]
    \centering
    \includegraphics[width=0.6\linewidth]{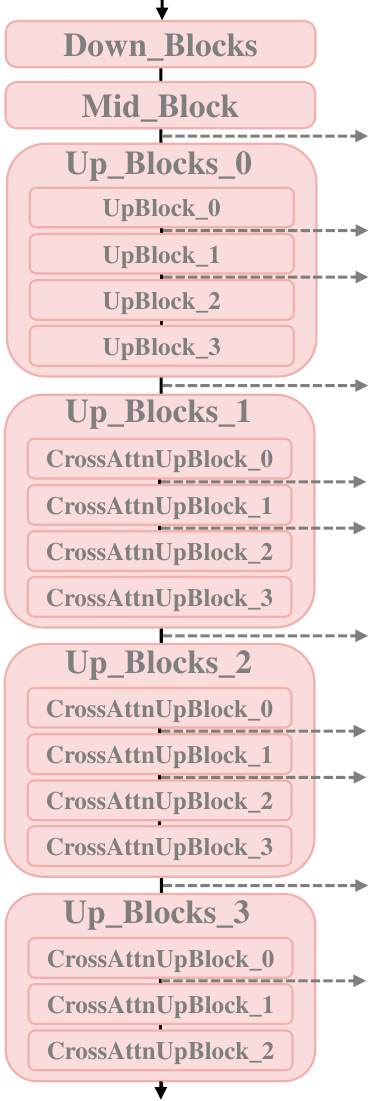}
    \caption{We carefully selected the internal representations from the U-Net-based D³-Predictor for self-supervised alignment.}
    \label{fig:arch_details}
    \vspace{-0.8em}
\end{figure}

\begin{figure}[t]
    \centering
    \includegraphics[width=0.6\linewidth]{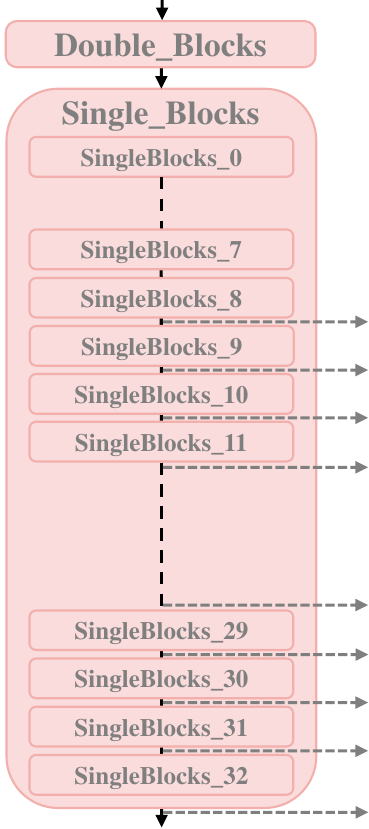}
    \caption{We carefully selected the internal representations from the DiT-based D³-Predictor for self-supervised alignment.}
    \label{fig:dit_arch}
    \vspace{-0.8em}
\end{figure}

\section{More Implementation Details}
\label{sec:More Implementation Details}
\subsection{Model Architecture}
\subsubsection{U-Net-Based D³-Predictor}
The U-Net-based D³-Predictor is initialized using the pretrained Stable Diffusion v2.1 (SD v2.1)~\cite{rombach2022high}. Figure~\ref{fig:arch_details} illustrates the backbone architecture of SD v2.1 and highlights the specific internal representations selected for self-supervised alignment (indicated by dashed arrows).

The selection of layers for representation alignment is critical. Prior work~\cite{tumanyan2023plug} suggests that $\text{SD}$'s upsampling blocks encode rich spatial semantic information that is largely invariant to appearance features (e.g., style and texture), focusing instead on underlying structure and shape. Inspired by this finding, we hypothesize that the internal representations from $\text{SD}$'s upsampling blocks are optimally suited for dense prediction, given that this task is fundamentally an image-to-structure mapping \cite{fu2024geowizard}. To maintain efficiency without compromising alignment quality, we carefully chose a minimal set of layers. This selection captures sufficient multi-scale structural detail while avoiding the high computational cost of aligning redundant representations. Consequently, we select the specific internal representations indicated by the dashed arrows in Figure~\ref{fig:arch_details}.

\subsubsection{DiT-Based D³-Predictor}
The DiT-based  D³-Predictor is initialized using the pretrained FLUX.1 dev model~\cite{flux2024}. Figure \ref{fig:dit_arch} presents the backbone architecture of FLUX.1 dev and indicates the internal representations selected for the self-supervised alignment (marked by dashed arrows). Drawing upon the insights from the U-Net-based D³-Predictor implementation, we focus on the representations from the deeper layers which typically carry refined spatial and structural information in DiT architectures. Specifically, we select the representations from the last $25$ single-flow blocks for robust and effective representation alignment.

\subsection{Text Prompt}
Text prompts provide diverse guidance for creative image \cite{li2024photomaker, xu2024diffusion, qin2023unicontrol} and video \cite{yang2024cogvideox, zhao2024moviedreamer, wang2024framer} generation, ensuring the results align with user intent. Despite their utility in creative tasks, extensive prior works \cite{ke2024repurposing, xu2024matters, zhao2025diception, ke2025marigold} suggest that complex text prompts are unnecessary for diffusion-based dense prediction models. This is primarily because dense prediction is a deterministic mapping from image to geometric structure~\cite{fu2024geowizard}, which does not require textual guidance for diversity. Consequently, the D³-Predictor utilizes a simple, fixed text prompt for each task to efficiently maintain the necessary context:
\begin{tcolorbox}[
    title={Text prompt for depth estimation},
    colback=gray!20,
    colframe=gray!90,
    ]
    \small
    A grayscale depth estimation image, where darker areas represent closer depths and lighter areas indicate farther depths.
\end{tcolorbox}
\begin{tcolorbox}[
    title={Text prompt for surface normal estimation},
    colback=gray!20,
    colframe=gray!90,
    ]
    \small
    A surface normal estimation map.
\end{tcolorbox}
\begin{tcolorbox}[
    title={Text prompt for image matting},
    colback=gray!20,
    colframe=gray!90,
    ]
    \small
    A human portrait matting map.
\end{tcolorbox}

\subsection{Training Datasets Mixing Strategy}
As described in Section $4.1.1$ of our main paper, we utilize multiple training datasets for both depth and surface normal estimation. Inspired by previous work \cite{garcia2025fine, fu2024geowizard}, we implement the following dataset sampling configuration in each iteration: For depth estimation, we sample Hypersim \cite{roberts2021hypersim} at a rate of 0.76, Virtual KITTI \cite{cabon2020virtual} at 0.12, and COCO \cite{lin2014microsoft} with pseudo-depth labels generated by Depth Pro \cite{bochkovskii2024depth} at 0.12; for surface normal estimation, we sample Hypersim \cite{roberts2021hypersim} at a rate of 0.9 and Sintel \cite{butler2012naturalistic} at 0.1.

\subsection{Training Objective}
As described in Section $3.2$ of our main paper, the training objective of the D³-Predictor is composed of two components that are jointly optimized. 

The first part of the objective is the \textit{Self-Supervised Diffusion Priors Aggregation}, which is implemented via the alignment of internal representations. Given that prior works \cite{yu2024representation, wu2025representation} have demonstrated the established efficacy of negative cosine similarity for representation alignment in diffusion models, we adopt it as the alignment metric $\text{dist}(\cdot,\cdot)$ in Equation $2$.

The second part of the objective is the \textit{Task-Specific Supervision}. We achieve this by minimizing the task-specific loss $\mathcal{L}_{\text{task}}$, which applies explicit supervision to the model's final prediction. 

For depth estimation, $\mathcal{L}_{\text{task}}$ is formulated as a weighted combination of three components: the MSE loss $\mathcal{L}_{\text{MSE}}$, the affine-invariant loss $\mathcal{L}_{\text{aff}}~$\cite{ranftl2020towards}, and the gradient loss in log space $\mathcal{L}_{\text{grad}}~$\cite{ranftl2020towards}. Specifically, $\mathcal{L}_{\text{task}}$ is defined as:
\begin{equation}
\mathcal{L}_{\text{task}} = \lambda_{\text{MSE}} \cdot \mathcal{L}_{\text{MSE}} + \lambda_{\text{aff}} \cdot \mathcal{L}_{\text{aff}} + \lambda_{\text{grad}} \cdot \mathcal{L}_{\text{grad}}
\end{equation}
 where the weighting coefficients are set to $\lambda_{\text{MSE}} = 8.0$, $\lambda_{\text{aff}} = 2.0$, and $\lambda_{\text{grad}} = 100.0$.

 For surface normal estimation, $\mathcal{L}_{\text{task}}$ is defined as a weighted sum of the MSE loss $\mathcal{L}_{\text{MSE}}$ and the angular loss $\mathcal{L}_{\text{ang}}$. Specifically, $\mathcal{L}_{\text{task}}$ is formulated as:
\begin{equation}
\mathcal{L}_{\text{task}} = \lambda_{\text{MSE}} \cdot \mathcal{L}_{\text{MSE}} + \lambda_{\text{ang}} \cdot \mathcal{L}_{\text{ang}}
\end{equation}
where the weighting coefficients are set to $\lambda_{\text{MSE}} = 8.0$ and $\lambda_{\text{ang}} = 3.0$.

For image matting, $\mathcal{L}_{\text{task}}$ is defined as a weighted sum of the MSE loss $\mathcal{L}_{\text{MSE}}$, the $\mathcal{L}_1$ loss $\mathcal{L}_{1}$, and the gradient loss $\mathcal{L}_{\text{grad}}$. Specifically, $\mathcal{L}_{\text{task}}$ is formulated as:
\begin{equation}
\mathcal{L}_{\text{task}} = \lambda_{\text{MSE}} \cdot \mathcal{L}_{\text{MSE}} + \lambda_{1} \cdot \mathcal{L}_{1} + \lambda_{\text{grad}} \cdot \mathcal{L}_{\text{grad}}
\end{equation}
where the weighting coefficients are set to $\lambda_{\text{MSE}} = 5.0$, $\lambda_{1} = 10.0$, and $\lambda_{\text{grad}} = 50.0$.

\section{Additional Visualization Results}
\label{sec:Additional Visualization Results}
\begin{figure*}[t]
    \centering
    \includegraphics[width=0.78\linewidth]{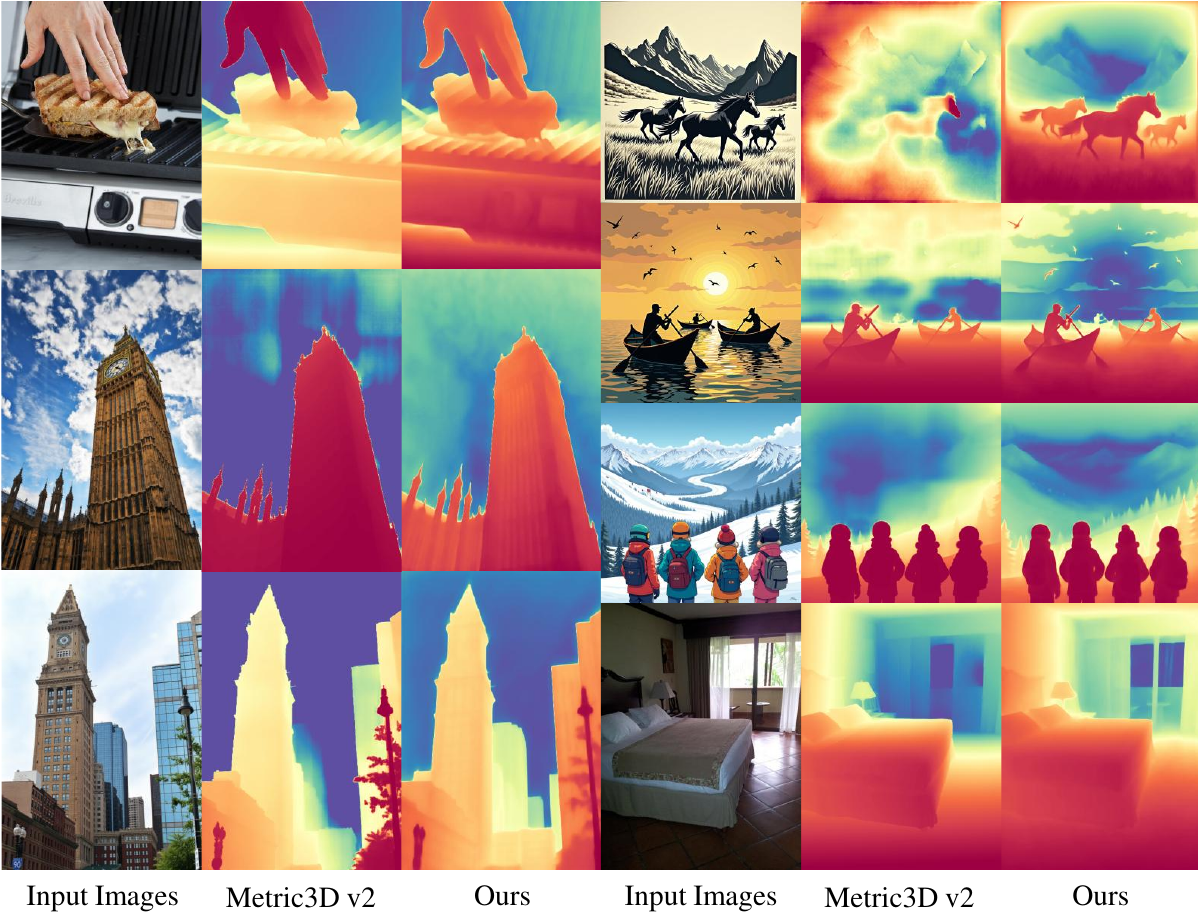}
    \vspace{-0.4em}
    \caption{More visualization results for the depth estimation task.}
    \label{fig:sup_depth}
    \vspace{-0.7em}
\end{figure*}

\begin{figure*}[t]
    \centering
    \includegraphics[width=0.78\linewidth]{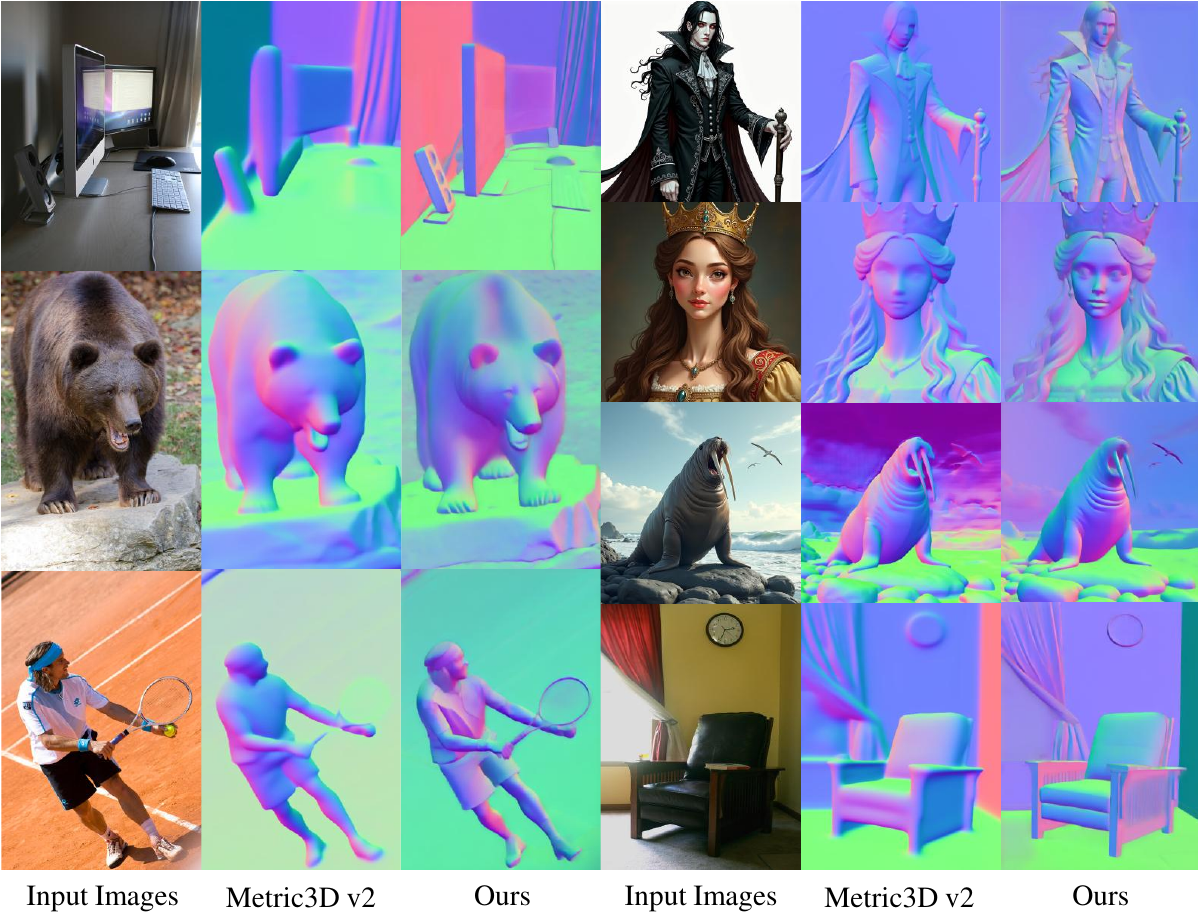}
    \vspace{-0.4em}
    \caption{More visualization results for the surface normal estimation task.}
    \label{fig:sup_normal}
\end{figure*}

\begin{figure*}[t]
    \centering
    \includegraphics[width=0.85\linewidth]{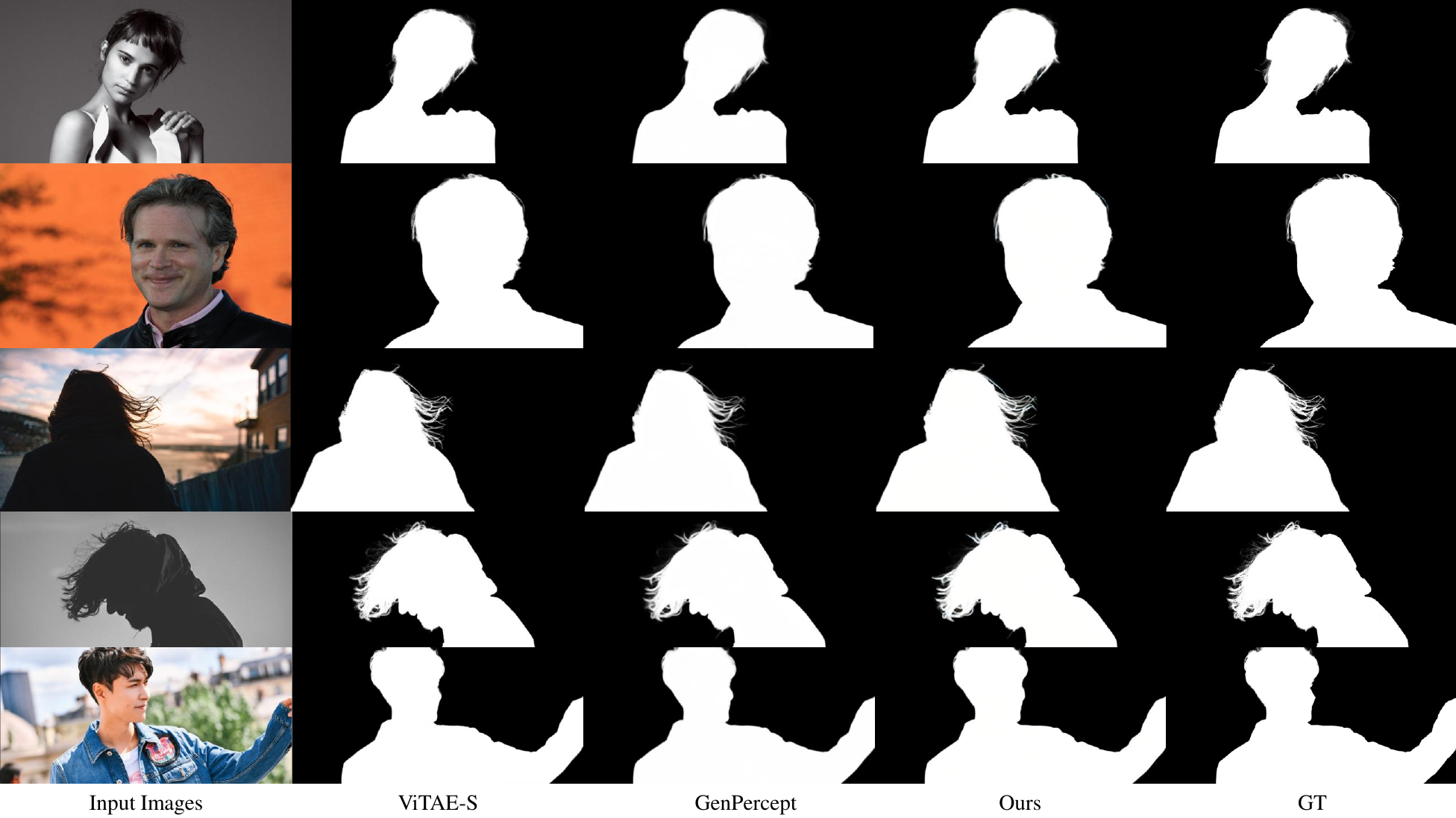}
    \vspace{-0.4em}
    \caption{More visualization results for the image matting task.}
    \label{fig:sup_matting}
    \vspace{-0.8em}
\end{figure*}

\begin{figure*}[t]
    \centering
    \includegraphics[width=0.85\linewidth]{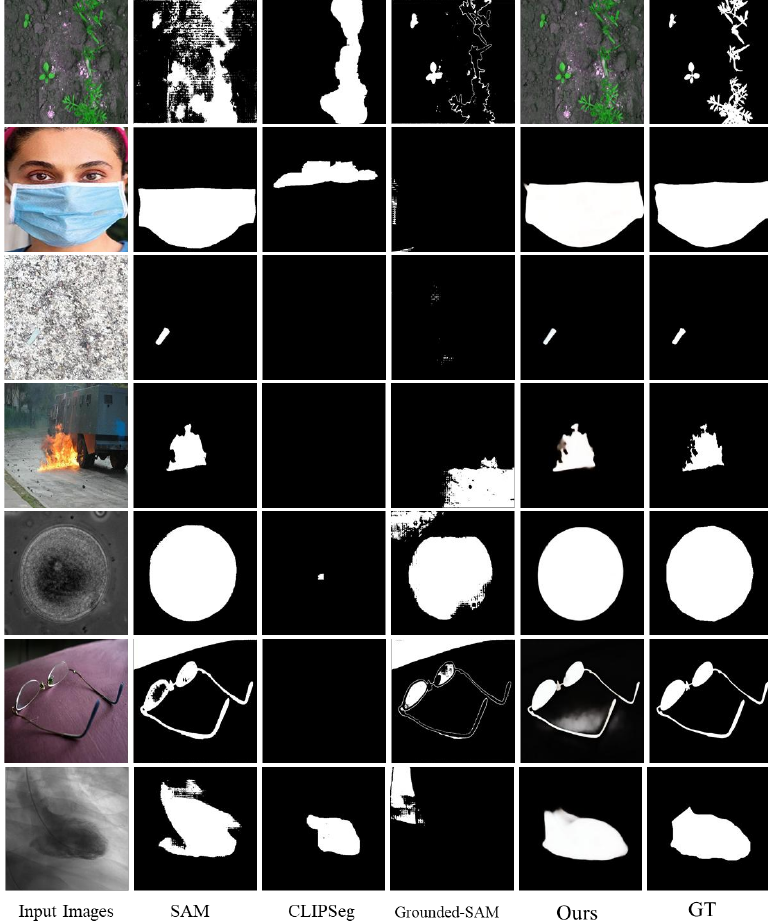}
    \caption{More visualization results for numerous practical downstream applications.}
    \label{fig:sup_real}
\end{figure*}

We provide additional visualization results in Figures \ref{fig:sup_depth} through \ref{fig:sup_real}, encompassing a diverse range of scenarios such as real-world scenes, $\text{AIGC}$ (AI-Generated Content) scenes, and domain-specific scenes. These comprehensive visualizations further demonstrate the $\text{D}^3$-Predictor's outstanding ability to generate fine geometric details, its strong generalization capability, and its practical utility.

\section{Additional Experimental Results}
\label{sec:Additional Experimental Results}

\subsection{Ablation on Projection Head}
We conduct an ablation study to investigate the contribution of the timestep-conditioned projection head used to map the internal representations of the D³-Predictor. We evaluate the performance of the D³-Predictor when trained with and without the projection head. Following the setup detailed in Section $4.3$ of our main paper, we report results on two standard depth estimation benchmarks: $\text{NYUv2}$ \cite{silberman2012indoor} for indoor scenes and $\text{KITTI}$ \cite{geiger2013vision} for driving scenes. The comparison results, shown in Table \ref{tab:ablation_projection_head}, consistently demonstrate that the D³-Predictor trained with the projection head outperforms the model trained without it on both benchmarks. This indicates that the projection head is essential for effectively bridging the representation of the D³-Predictor and the timestep-dependent representations extracted from the pretrained diffusion model.

\begin{table}[h]
    \centering
    \caption{Ablation study on the projection head. Other training settings are the same as our main experiments.}
    \label{tab:ablation_projection_head}
    \resizebox{0.95\columnwidth}{!}{%
      \begin{tabular}{M{3.0cm}*{4}{M{1.4cm}}}
        \toprule
        \multirow{2}{*}{Projection Head} & \multicolumn{2}{c}{KITTI} & \multicolumn{2}{c}{NYUv2} \\
        \cmidrule(lr){2-3}\cmidrule(lr){4-5}
         & AbsRel$\downarrow$ & $\delta_1\uparrow$ & AbsRel$\downarrow$ & $\delta_1\uparrow$ \\
        \midrule
        Yes & 0.082 & 0.940 & 0.052 & 0.970 \\
        No     & 0.089 & 0.930 & 0.057 & 0.964 \\
        \bottomrule
      \end{tabular}%
    }
\end{table}

\subsection{Necessity of Synthetic Training Data}
In our main experiments, the optimal performance of the D³-Predictor is achieved by training with a mix of synthetic data and pseudo-labeled real data. During training, synthetic data provides high-quality geometric details, while real data contributes to diverse-scene generalization. Section $4.3.2$ empirically demonstrates the generalization benefits introduced by real images. Meanwhile, we argue that high-quality synthetic data is also indispensable for strong model performance, as evidenced by the results in Table~\ref{tab:necessity_of_synthetic_training_data}.

\begin{table}[h]
    \centering
    \caption{Performance comparison of D³-Predictor under different training data configurations.}
    \label{tab:necessity_of_synthetic_training_data}
    \resizebox{0.95\columnwidth}{!}{%
      \begin{tabular}{M{3.0cm}*{4}{M{1.4cm}}}
        \toprule
        \multirow{2}{*}{Synthetic + Real} & \multicolumn{2}{c}{KITTI} & \multicolumn{2}{c}{NYUv2} \\
        \cmidrule(lr){2-3}\cmidrule(lr){4-5}
         & AbsRel$\downarrow$ & $\delta_1\uparrow$ & AbsRel$\downarrow$ & $\delta_1\uparrow$ \\
        \midrule
        7.5K + 7.5K & 0.087 & 0.933 & 0.060 & 0.967 \\
        15K + 0     & 0.092 & 0.923 & 0.061 & 0.963 \\
        0 + 15K    & 0.089 & 0.928 & 0.061 & 0.960 \\
        \bottomrule
      \end{tabular}%
    }
\end{table}

\subsection{Efficiency and Runtime Analysis}
We conduct a thorough analysis of the efficiency of the D³-Predictor, summarizing the results in Table $\text{\ref{tab:runtime}}$. The reported inference time is the average runtime over $5,000$ images at $768 \times 768$ resolution on a single NVIDIA $\text{L40S}$ $\text{GPU}$. Our D³-Predictor achieves the fastest inference speed among all evaluated baselines. Specifically, our method is approximately $66\%$ faster than the discriminative model Metric3Dv2~\cite{hu2024metric3d} and achieves nearly a $10\%$ speedup over the fastest comparable diffusion-based methods, such as E2E-FT~\cite{garcia2025fine} and Lotus-G~\cite{he2024lotus}.

Beyond speed, the D³-Predictor demonstrates highly competitive $\text{GPU}$ memory efficiency. While $\text{Lotus-G}$~\cite{he2024lotus} maintains the minimum memory consumption, our model's usage is comparable and represents a significant saving of nearly $50\%$ compared to the most memory-intensive baseline, Depth Pro~\cite{bochkovskii2024depth}. Even when compared to other efficient diffusion-based models like E2E-FT~\cite{garcia2025fine}, we still achieve over $20\%$ savings in memory consumption. This excellent performance in both inference time and memory requirement confirms that the D³-Predictor is an efficiency diffusion-based method while maintaining high predictive performance.

\begin{table}[htbp]
    \centering
    \small
    \setlength{\tabcolsep}{10pt}
    \caption{Efficiency and runtime analysis. We compare D³-Predictor with two leading discriminative models and selected diffusion models operating in \textit{NFEs=1} mode. Diffusion-based methods that require ensemble procedures or multi-step inference are excluded due to their significantly higher computational costs.}
    \label{tab:runtime}
    \begin{tabular}{lcc}
        \toprule
        \textbf{Method} & \textbf{Time (s)} & \textbf{Memory (GB)} \\
        \midrule
        \multicolumn{3}{l}{\textit{Discriminative Models}} \\
        Metric3Dv2~\cite{hu2024metric3d} & $\sim 0.66$ & $7.5$ \\
        Depth Pro~\cite{bochkovskii2024depth} & $\sim 0.64$ & $12.2$ \\
        \midrule
        \multicolumn{3}{l}{\textit{Diffusion Models (NFEs=1)}} \\
        E2E-FT~\cite{garcia2025fine} & $\sim 0.27$ & $8.07$ \\
        Marigold-Normals~\cite{ke2025marigold} & $\sim 0.45$ & $8.04$ \\
        Lotus-G~\cite{he2024lotus} & $\sim 0.25$ & $5.72$ \\
        GenPercept~\cite{xu2024matters} & $\sim 0.28$ & $8.05$ \\
        \midrule
        Ours & $\sim 0.22$ & 6.38 \\
        \bottomrule
    \end{tabular}
\end{table}

\subsection{Generalization to Domain-Specific Tasks}
Section $4.2.4$ of our main paper further evaluates the generalization capability and practical utility of the D³-Predictor across $20$ additional domain-specific dense prediction tasks. These tasks cover a wide range of practical application fields, including healthcare, ecological monitoring, public safety, and infrastructure inspection. A comprehensive overview of the results on these $20$ tasks is presented in Tables \ref{tab:pavement_crack_detect} through \ref{tab:cardiac_segmentation}. Our model significantly outperforms all zero-shot baselines across all $20$ tasks, demonstrating strong zero-shot generalization and establishing a robust foundation model for numerous practical downstream applications.

\begin{table}[htbp]
    \centering
    \scriptsize
    \setlength{\tabcolsep}{6pt}
    \renewcommand{\arraystretch}{1.2}
    \caption{Evaluation of the \textit{Pavement Crack Detect} task~\cite{zhang2016road}.}
    \label{tab:pavement_crack_detect}
    \begin{tabular}{lccccc}
    \toprule
    Model & \makecell{Training\\ Samples} & IoU $\uparrow$ & PA $\uparrow$ & DiCE $\uparrow$  \\
    \midrule
    SAM~\cite{kirillov2023segment} & 1.1B & 0.351 & 0.453 & 0.400  \\
    CLIPSeg~\cite{luddecke2022image} & 345K & \underline{0.487} & \underline{0.499} & \underline{0.493}  \\
    Grounded-SAM~\cite{ren2024grounded} & 1.1B & 0.393 & 0.429 & 0.428  \\
    \rowcolor{gray!10}
    Ours & 15 & \textbf{0.694} & \textbf{0.730} & \textbf{0.766}  \\
    \bottomrule
    \end{tabular}
\end{table}

\begin{table}[htbp]
    \centering
    \scriptsize
    \setlength{\tabcolsep}{6pt}
    \renewcommand{\arraystretch}{1.2}
    \caption{Evaluation of the \textit{Auto. Pothole Detect} task~\cite{fan2019pothole}.}
    \label{tab:auto_pothole_detect}
    \begin{tabular}{lccccc}
    \toprule
    Model & \makecell{Training\\ Samples} & IoU $\uparrow$ & PA $\uparrow$ & DiCE $\uparrow$  \\
    \midrule
    SAM~\cite{kirillov2023segment} & 1.1B & 0.162 & 0.415 & 0.186  \\
    CLIPSeg~\cite{luddecke2022image} & 345K & \underline{0.597} & \underline{0.819} & \underline{0.668}  \\
    Grounded-SAM~\cite{ren2024grounded} & 1.1B & 0.429 & 0.469 & 0.460  \\
    \rowcolor{gray!10}
    Ours & 15 & \textbf{0.793} & \textbf{0.828} & \textbf{0.839}  \\
    \bottomrule
    \end{tabular}
\end{table}

\begin{table}[htbp]
    \centering
    \scriptsize
    \setlength{\tabcolsep}{6pt}
    \renewcommand{\arraystretch}{1.2}
    \caption{Evaluation of the \textit{Pedestrian Safety Monitor} task~\cite{wang2007object}.}
    \label{tab:pedestrian_safety_monitor}
    \begin{tabular}{lccccc}
    \toprule
    Model & \makecell{Training\\ Samples} & IoU $\uparrow$ & PA $\uparrow$ & DiCE $\uparrow$  \\
    \midrule
    SAM~\cite{kirillov2023segment} & 1.1B & 0.417 & 0.495 & 0.478  \\
    CLIPSeg~\cite{luddecke2022image} & 345K & \underline{0.616} & \underline{0.717} & \underline{0.687}  \\
    Grounded-SAM~\cite{ren2024grounded} & 1.1B & 0.423 & 0.505 & 0.478  \\
    \rowcolor{gray!10}
    Ours & 15 & \textbf{0.878} & \textbf{0.939} & \textbf{0.929}  \\
    \bottomrule
    \end{tabular}
\end{table}

\begin{table}[htbp]
    \centering
    \scriptsize
    \setlength{\tabcolsep}{6pt}
    \renewcommand{\arraystretch}{1.2}
    \caption{Evaluation of the \textit{Precise Road Model} task~\cite{mnih2013machine}.}
    \label{tab:precise_road_model}
    \begin{tabular}{lccccc}
    \toprule
    Model & \makecell{Training\\ Samples} & IoU $\uparrow$ & PA $\uparrow$ & DiCE $\uparrow$  \\
    \midrule
    SAM~\cite{kirillov2023segment} & 1.1B & 0.434 & 0.465 & 0.464  \\
    CLIPSeg~\cite{luddecke2022image} & 345K & \underline{0.439} & \underline{0.562} & \underline{0.505}  \\
    Grounded-SAM~\cite{ren2024grounded} & 1.1B & 0.278 & 0.472 & 0.357  \\
    \rowcolor{gray!10}
    Ours & 15 & \textbf{0.646} & \textbf{0.711} & \textbf{0.731}  \\
    \bottomrule
    \end{tabular}
\end{table}

\begin{table}[htbp]
    \centering
    \scriptsize
    \setlength{\tabcolsep}{6pt}
    \renewcommand{\arraystretch}{1.2}
    \caption{Evaluation of the \textit{Retinal Vessel Analysis} task~\cite{jin2022fives}.}
    \label{tab:retinal_vessel_analysis}
    \begin{tabular}{lccccc}
    \toprule
    Model & \makecell{Training\\ Samples} & IoU $\uparrow$ & PA $\uparrow$ & DiCE $\uparrow$  \\
    \midrule
    SAM~\cite{kirillov2023segment} & 1.1B & 0.273 & \underline{0.594} & 0.370  \\
    CLIPSeg~\cite{luddecke2022image} & 345K & \underline{0.472} & 0.510 & \underline{0.499}  \\
    Grounded-SAM~\cite{ren2024grounded} & 1.1B & 0.344 & 0.376 & 0.409  \\
    \rowcolor{gray!10}
    Ours & 15 & \textbf{0.757} & \textbf{0.789} & \textbf{0.837}  \\
    \bottomrule
    \end{tabular}
\end{table}

\begin{table}[htbp]
    \centering
    \scriptsize
    \setlength{\tabcolsep}{6pt}
    \renewcommand{\arraystretch}{1.2}
    \caption{Evaluation of the \textit{Spinal Morphology Assessment} task~\cite{chu2015annotated}.}
    \label{tab:spinal_morphology_assessment}
    \begin{tabular}{lccccc}
    \toprule
    Model & \makecell{Training\\ Samples} & IoU $\uparrow$ & PA $\uparrow$ & DiCE $\uparrow$  \\
    \midrule
    SAM~\cite{kirillov2023segment} & 1.1B & \underline{0.555} & \underline{0.611} & \underline{0.612}  \\
    CLIPSeg~\cite{luddecke2022image} & 345K & 0.519 & 0.569 & 0.551  \\
    Grounded-SAM~\cite{ren2024grounded} & 1.1B & 0.464 & 0.481 & 0.481  \\
    \rowcolor{gray!10}
    Ours & 15 & \textbf{0.932} & \textbf{0.969} & \textbf{0.963}  \\
    \bottomrule
    \end{tabular}
\end{table}

\begin{table}[htbp]
    \centering
    \scriptsize
    \setlength{\tabcolsep}{6pt}
    \renewcommand{\arraystretch}{1.2}
    \caption{Evaluation of the \textit{Oocyte Detect} task~\cite{letort2022interpretable}.}
    \label{tab:oocyte_detect}
    \begin{tabular}{lccccc}
    \toprule
    Model & \makecell{Training\\ Samples} & IoU $\uparrow$ & PA $\uparrow$ & DiCE $\uparrow$  \\
    \midrule
    SAM~\cite{kirillov2023segment} & 1.1B & \underline{0.952} & \underline{0.974} & \underline{0.969}  \\
    CLIPSeg~\cite{luddecke2022image} & 345K & 0.366 & 0.500 & 0.422  \\
    Grounded-SAM~\cite{ren2024grounded} & 1.1B & 0.390 & 0.518 & 0.460  \\
    \rowcolor{gray!10}
    Ours & 15 & \textbf{0.981} & \textbf{0.991} & \textbf{0.990}  \\
    \bottomrule
    \end{tabular}
\end{table}

\begin{table}[htbp]
    \centering
    \scriptsize
    \setlength{\tabcolsep}{6pt}
    \renewcommand{\arraystretch}{1.2}
    \caption{Evaluation of the \textit{Nucleus Localization} task~\cite{naylor2018segmentation}.}
    \label{tab:nucleus_localization}
    \begin{tabular}{lccccc}
    \toprule
    Model & \makecell{Training\\ Samples} & IoU $\uparrow$ & PA $\uparrow$ & DiCE $\uparrow$  \\
    \midrule
    SAM~\cite{kirillov2023segment} & 1.1B & 0.277 & 0.498 & 0.336  \\
    CLIPSeg~\cite{luddecke2022image} & 345K & \underline{0.452} & \underline{0.511}& 0.488  \\
    Grounded-SAM~\cite{ren2024grounded} & 1.1B & 0.447 & 0.506 & \underline{0.496}  \\
    \rowcolor{gray!10}
    Ours & 15 & \textbf{0.814} & \textbf{0.880} & \textbf{0.884}  \\
    \bottomrule
    \end{tabular}
\end{table}

\begin{table}[htbp]
    \centering
    \scriptsize
    \setlength{\tabcolsep}{6pt}
    \renewcommand{\arraystretch}{1.2}
    \caption{Evaluation of the \textit{Water Body Map} task~\cite{2020SAR}.}
    \label{tab:water_body_map}
    \begin{tabular}{lccccc}
    \toprule
    Model & \makecell{Training\\ Samples} & IoU $\uparrow$ & PA $\uparrow$ & DiCE $\uparrow$  \\
    \midrule
    SAM~\cite{kirillov2023segment} & 1.1B & 0.299 & 0.415 & 0.348  \\
    CLIPSeg~\cite{luddecke2022image} & 345K & 0.172 & \underline{0.504} & 0.235  \\
    Grounded-SAM~\cite{ren2024grounded} & 1.1B & \underline{0.322} & 0.435 & \underline{0.406}  \\
    \rowcolor{gray!10}
    Ours & 15 & \textbf{0.690} & \textbf{0.863} & \textbf{0.754}  \\
    \bottomrule
    \end{tabular}
\end{table}

\begin{table}[htbp]
    \centering
    \scriptsize
    \setlength{\tabcolsep}{6pt}
    \renewcommand{\arraystretch}{1.2}
    \caption{Evaluation of the \textit{Auto. Tree Tagging} task~\cite{Tree-Binary-Segmentation}.}
    \label{tab:auto_tree_tagging}
    \begin{tabular}{lccccc}
    \toprule
    Model & \makecell{Training\\ Samples} & IoU $\uparrow$ & PA $\uparrow$ & DiCE $\uparrow$  \\
    \midrule
    SAM~\cite{kirillov2023segment} & 1.1B & \underline{0.765} & \underline{0.840} & \underline{0.809}  \\
    CLIPSeg~\cite{luddecke2022image} & 345K & 0.463 & 0.722 & 0.579  \\
    Grounded-SAM~\cite{ren2024grounded} & 1.1B & 0.391 & 0.529 & 0.457  \\
    \rowcolor{gray!10}
    Ours & 15 & \textbf{0.857} & \textbf{0.931} & \textbf{0.898}  \\
    \bottomrule
    \end{tabular}
\end{table}

\begin{table}[htbp]
    \centering
    \scriptsize
    \setlength{\tabcolsep}{6pt}
    \renewcommand{\arraystretch}{1.2}
    \caption{Evaluation of the \textit{Camouflage Detect} task~\cite{fan2020camouflaged}.}
    \label{tab:camouflage_detect}
    \begin{tabular}{lccccc}
    \toprule
    Model & \makecell{Training\\ Samples} & IoU $\uparrow$ & PA $\uparrow$ & DiCE $\uparrow$  \\
    \midrule
    SAM~\cite{kirillov2023segment} & 1.1B & 0.548 & 0.612 & 0.601  \\
    CLIPSeg~\cite{luddecke2022image} & 345K & \underline{0.569} & \underline{0.753} & \underline{0.667}  \\
    Grounded-SAM~\cite{ren2024grounded} & 1.1B & 0.454 & 0.504 & 0.497  \\
    \rowcolor{gray!10}
    Ours & 15 & \textbf{0.714} & \textbf{0.784} & \textbf{0.770}  \\
    \bottomrule
    \end{tabular}
\end{table}

\begin{table}[htbp]
    \centering
    \scriptsize
    \setlength{\tabcolsep}{6pt}
    \renewcommand{\arraystretch}{1.2}
    \caption{Evaluation of the \textit{Oil Spill Track} task~\cite{zhu2021oil}.}
    \label{tab:oil_spill_track}
    \begin{tabular}{lccccc}
    \toprule
    Model & \makecell{Training\\ Samples} & IoU $\uparrow$ & PA $\uparrow$ & DiCE $\uparrow$  \\
    \midrule
    SAM~\cite{kirillov2023segment} & 1.1B & 0.133 & 0.437 & 0.187  \\
    CLIPSeg~\cite{luddecke2022image} & 345K & \underline{0.427} & \underline{0.618} & \underline{0.511}  \\
    Grounded-SAM~\cite{ren2024grounded} & 1.1B & 0.376 & 0.490 & 0.463  \\
    \rowcolor{gray!10}
    Ours & 15 & \textbf{0.761} & \textbf{0.860} & \textbf{0.839}  \\
    \bottomrule
    \end{tabular}
\end{table}

\begin{table}[htbp]
    \centering
    \scriptsize
    \setlength{\tabcolsep}{6pt}
    \renewcommand{\arraystretch}{1.2}
    \caption{Evaluation of the \textit{Urban Layout Analysis} task~\cite{mnih2013machine}.}
    \label{tab:urban_layout_analysis}
    \begin{tabular}{lccccc}
    \toprule
    Model & \makecell{Training\\ Samples} & IoU $\uparrow$ & PA $\uparrow$ & DiCE $\uparrow$  \\
    \midrule
    SAM~\cite{kirillov2023segment} & 1.1B & \underline{0.386} & 0.475 & \underline{0.437}  \\
    CLIPSeg~\cite{luddecke2022image} & 345K & 0.315 & \underline{0.616} & 0.421  \\
    Grounded-SAM~\cite{ren2024grounded} & 1.1B & 0.284 & 0.555 & 0.390  \\
    \rowcolor{gray!10}
    Ours & 15 & \textbf{0.653} & \textbf{0.741} & \textbf{0.756}  \\
    \bottomrule
    \end{tabular}
\end{table}

\begin{table}[htbp]
    \centering
    \scriptsize
    \setlength{\tabcolsep}{6pt}
    \renewcommand{\arraystretch}{1.2}
    \caption{Evaluation of the \textit{Weed Distribution Mapping} task~\cite{haug15}.}
    \label{tab:weed_distrib_map}
    \begin{tabular}{lccccc}
    \toprule
    Model & \makecell{Training\\ Samples} & IoU $\uparrow$ & PA $\uparrow$ & DiCE $\uparrow$  \\
    \midrule
    SAM~\cite{kirillov2023segment} & 1.1B & 0.364 & 0.399 & 0.398  \\
    CLIPSeg~\cite{luddecke2022image} & 345K & \underline{0.609} & \underline{0.858} & \underline{0.725}  \\
    Grounded-SAM~\cite{ren2024grounded} & 1.1B & 0.126 & 0.420 & 0.173  \\
    \rowcolor{gray!10}
    Ours & 15 & \textbf{0.924} & \textbf{0.952} & \textbf{0.959}  \\
    \bottomrule
    \end{tabular}
\end{table}

\begin{table}[htbp]
    \centering
    \scriptsize
    \setlength{\tabcolsep}{6pt}
    \renewcommand{\arraystretch}{1.2}
    \caption{Evaluation of the \textit{Mask-Wear Monitor} task~\cite{rezvani2024abanet}.}
    \label{tab:mask_wear_monitor}
    \begin{tabular}{lccccc}
    \toprule
    Model & \makecell{Training\\ Samples} & IoU $\uparrow$ & PA $\uparrow$ & DiCE $\uparrow$  \\
    \midrule
    SAM~\cite{kirillov2023segment} & 1.1B & \underline{0.540} & \underline{0.614} & \underline{0.584}  \\
    CLIPSeg~\cite{luddecke2022image} & 345K & 0.377 & 0.513 & 0.469  \\
    Grounded-SAM~\cite{ren2024grounded} & 1.1B & 0.377 & 0.498 & 0.405  \\
    \rowcolor{gray!10}
    Ours & 15 & \textbf{0.903} & \textbf{0.943} & \textbf{0.942}  \\
    \bottomrule
    \end{tabular}
\end{table}

\begin{table}[htbp]
    \centering
    \scriptsize
    \setlength{\tabcolsep}{6pt}
    \renewcommand{\arraystretch}{1.2}
    \caption{Evaluation of the \textit{Cigarette Detect} task~\cite{cigarettebutts2024}.}
    \label{tab:cigarette_detect}
    \begin{tabular}{lccccc}
    \toprule
    Model & \makecell{Training\\ Samples} & IoU $\uparrow$ & PA $\uparrow$ & DiCE $\uparrow$  \\
    \midrule
    SAM~\cite{kirillov2023segment} & 1.1B & 0.304 & 0.308 & 0.324  \\
    CLIPSeg~\cite{luddecke2022image} & 345K & \underline{0.495} & \underline{0.500} & \underline{0.497}  \\
    Grounded-SAM~\cite{ren2024grounded} & 1.1B & 0.479 & 0.483 & 0.486  \\
    \rowcolor{gray!10}
    Ours & 15 & \textbf{0.875} & \textbf{0.908} & \textbf{0.904}  \\
    \bottomrule
    \end{tabular}
\end{table}

\begin{table}[htbp]
    \centering
    \scriptsize
    \setlength{\tabcolsep}{6pt}
    \renewcommand{\arraystretch}{1.2}
    \caption{Evaluation of the \textit{Fire Alert} task~\cite{chino2015bowfire}.}
    \label{tab:fire_alert_sys}
    \begin{tabular}{lccccc}
    \toprule
    Model & \makecell{Training\\ Samples} & IoU $\uparrow$ & PA $\uparrow$ & DiCE $\uparrow$  \\
    \midrule
    SAM~\cite{kirillov2023segment} & 1.1B & 0.528 & 0.591 & 0.574  \\
    CLIPSeg~\cite{luddecke2022image} & 345K & \underline{0.642} & \underline{0.753} & \underline{0.710}  \\
    Grounded-SAM~\cite{ren2024grounded} & 1.1B & 0.469 & 0.530 & 0.516  \\
    \rowcolor{gray!10}
    Ours & 15 & \textbf{0.829} & \textbf{0.951} & \textbf{0.889}  \\
    \bottomrule
    \end{tabular}
\end{table}

\begin{table}[htbp]
    \centering
    \scriptsize
    \setlength{\tabcolsep}{6pt}
    \renewcommand{\arraystretch}{1.2}
    \caption{Evaluation of the \textit{Food Cell Inspect} task~\cite{letort2022interpretable}.}
    \label{tab:food_cell_inspect}
    \begin{tabular}{lccccc}
    \toprule
    Model & \makecell{Training\\ Samples} & IoU $\uparrow$ & PA $\uparrow$ & DiCE $\uparrow$  \\
    \midrule
    SAM~\cite{kirillov2023segment} & 1.1B & \underline{0.936} & \underline{0.961} & \underline{0.959}  \\
    CLIPSeg~\cite{luddecke2022image} & 345K & 0.291 & 0.500 & 0.367  \\
    Grounded-SAM~\cite{ren2024grounded} & 1.1B & 0.402 & 0.561 & 0.487  \\
    \rowcolor{gray!10}
    Ours & 15 & \textbf{0.976} & \textbf{0.988} & \textbf{0.987}  \\
    \bottomrule
    \end{tabular}
\end{table}

\begin{table}[htbp]
    \centering
    \scriptsize
    \setlength{\tabcolsep}{6pt}
    \renewcommand{\arraystretch}{1.2}
    \caption{Evaluation of the \textit{Close-Range Segment} task~\cite{qin2022highly}.}
    \label{tab:close_range_segment}
    \begin{tabular}{lccccc}
    \toprule
    Model & \makecell{Training\\ Samples} & IoU $\uparrow$ & PA $\uparrow$ & DiCE $\uparrow$  \\
    \midrule
    SAM~\cite{kirillov2023segment} & 1.1B & 0.412 & 0.478 & 0.474  \\
    CLIPSeg~\cite{luddecke2022image} & 345K & 0.429 & 0.498 & 0.459  \\
    Grounded-SAM~\cite{ren2024grounded} & 1.1B & \underline{0.436} & \underline{0.566} & \underline{0.516}  \\
    \rowcolor{gray!10}
    Ours & 15 & \textbf{0.655} & \textbf{0.734} & \textbf{0.731}  \\
    \bottomrule
    \end{tabular}
\end{table}

\begin{table}[htbp]
    \centering
    \scriptsize
    \setlength{\tabcolsep}{6pt}
    \renewcommand{\arraystretch}{1.2}
    \caption{Evaluation of the \textit{Cardiac Screen} task~\cite{cardiacsegmentation2020}.}
    \label{tab:cardiac_segmentation}
    \begin{tabular}{lccccc}
    \toprule
    Model & \makecell{Training\\ Samples} & IoU $\uparrow$ & PA $\uparrow$ & DiCE $\uparrow$  \\
    \midrule
    SAM~\cite{kirillov2023segment} & 1.1B & \underline{0.535} & \underline{0.611} & \underline{0.612}  \\
    CLIPSeg~\cite{luddecke2022image} & 345K & 0.519 & 0.569 & 0.551  \\
    Grounded-SAM~\cite{ren2024grounded} & 1.1B & 0.464 & 0.481 & 0.481  \\
    \rowcolor{gray!10}
    Ours & 15 & \textbf{0.885} & \textbf{0.959} & \textbf{0.935}  \\
    \bottomrule
    \end{tabular}
\end{table}

\clearpage

\begin{figure*}[t]
    \centering
    \includegraphics[width=0.85\linewidth]{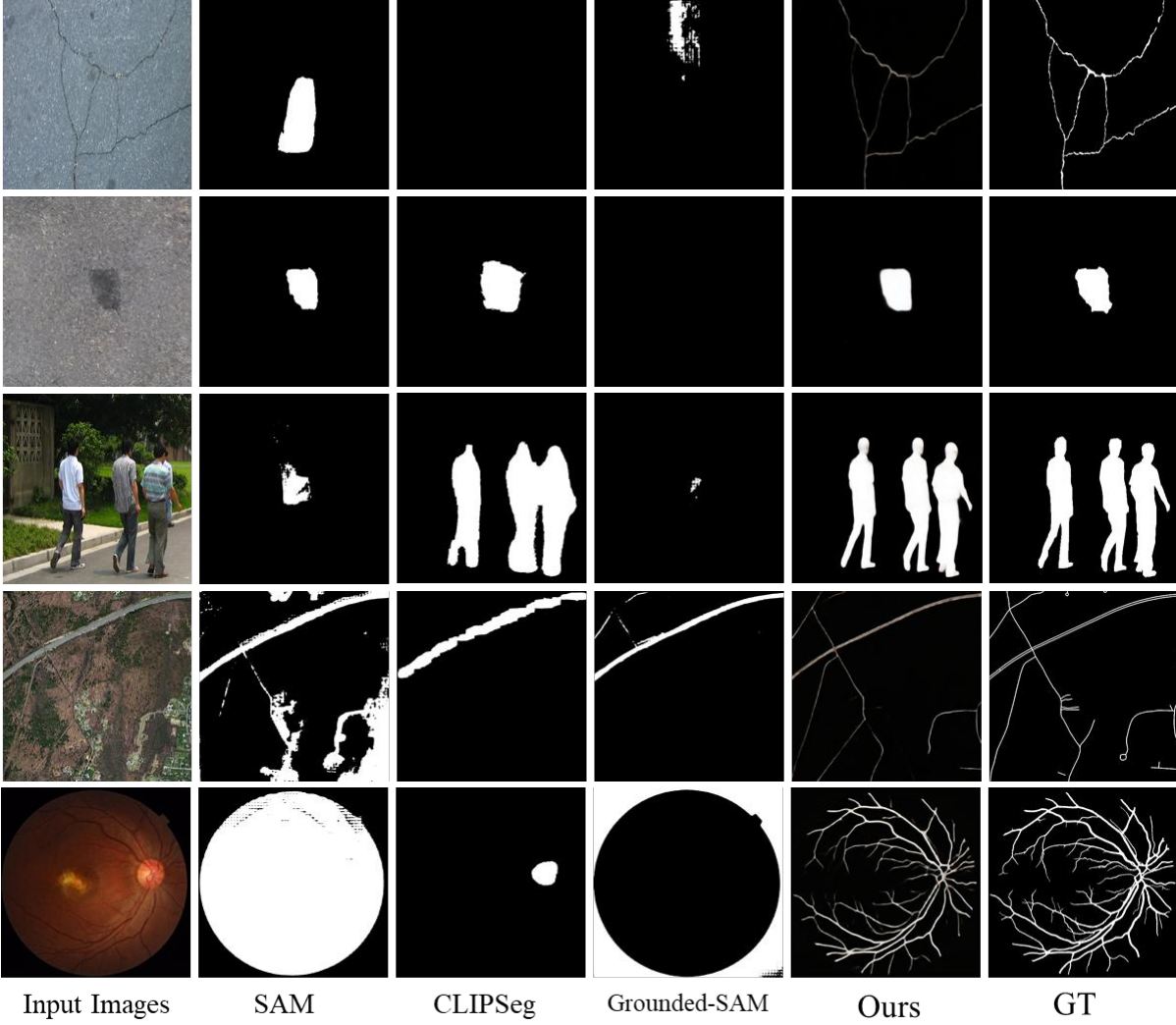}
    \vspace{-0.4em}
\end{figure*}

\begin{figure*}[t]
    \centering
    \includegraphics[width=0.85\linewidth]{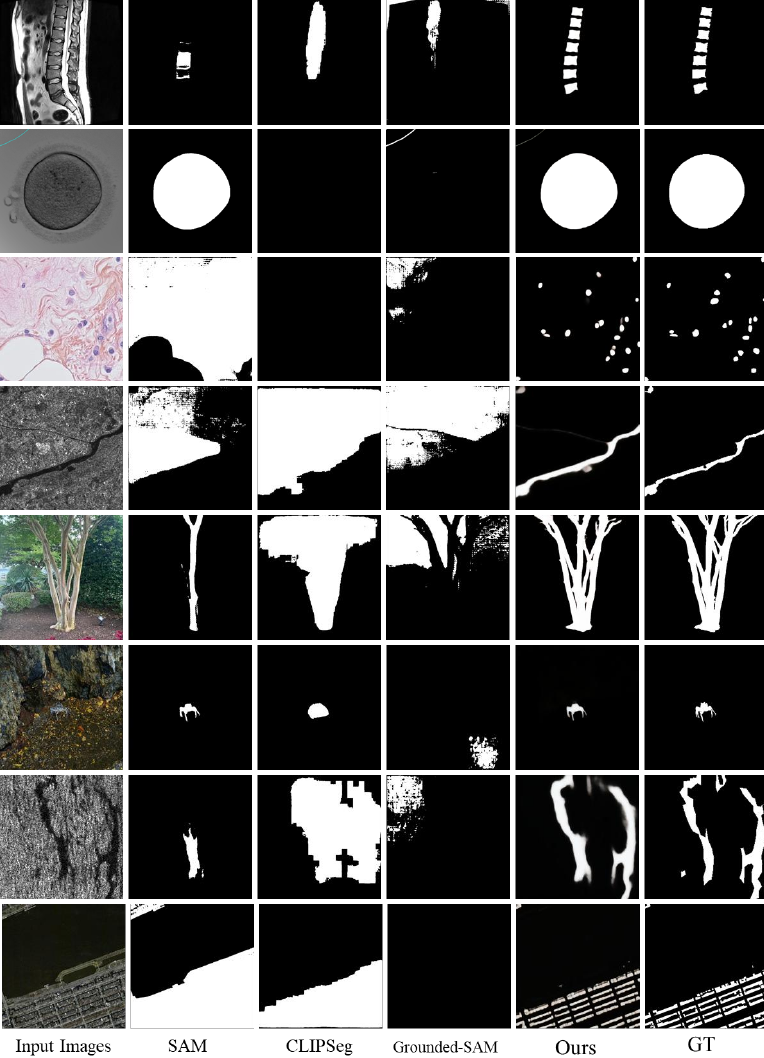}
\end{figure*}

